\documentclass[a4paper,11pt]{article}

\usepackage{publication}
\usepackage[backend=biber]{biblatex}
\title{Cycling Race Time Prediction: A Personalized Machine Learning Approach Using Route Topology and Training Load}

\author{Francisco Aguilera Moreno - BSc Computer Science}
\contact{aguimorefran@gmail.com}

\date{December 2025}

\begin{document}

\maketitle

\begin{abstract}
Predicting cycling duration for a given route is essential for training planning and event preparation.
Existing solutions rely on physics-based models that require extensive parameterization, including
aerodynamic drag coefficients and real-time wind forecasts, parameters impractical for most amateur
cyclists. This work presents a machine learning approach that predicts ride duration using route
topology features combined with the athlete's current fitness state derived from training load metrics.
The model learns athlete-specific performance patterns from historical data, substituting complex
physical measurements with historical performance proxies. We evaluate the approach using a
single-athlete dataset (N=96 rides) in an N-of-1 study design. After rigorous feature engineering
to eliminate data leakage, we find that Lasso regression with \textbf{Topology + Fitness} features
achieves MAE=6.60 minutes and R²=0.922. Notably, integrating fitness metrics (Chronic Training Load [CTL], Acute Training Load [ATL]) reduces
error by 14\% compared to topology alone (MAE=7.66 min), demonstrating that physiological state
meaningfully constrains performance even in self-paced efforts. Progressive checkpoint predictions
enable dynamic race planning as route difficulty becomes apparent.

\end{abstract}

\tableofcontents
\vspace{1em}

\section{Introduction}

\subsection{Motivation}

Cycling duration prediction is a practical problem faced by athletes at all levels. Whether planning
a training session, estimating arrival times for a group ride, or pacing a competitive event, knowing
how long a route will take provides valuable information for decision-making. This need becomes
particularly relevant when routes involve significant elevation changes, where naive distance-based
estimates fail to account for the time cost of climbing.

Current solutions to this problem fall into two categories. Physics-based tools, such as Best Bike
Split, solve equations of motion considering power output, aerodynamic drag, rolling resistance,
and environmental conditions. While accurate, these approaches require parameters that are difficult
to obtain: aerodynamic drag coefficients typically require wind tunnel testing or field measurements,
and accurate wind forecasts are rarely available for arbitrary routes. The alternative, simple
rule-of-thumb estimates based on average speed, ignores the significant impact of terrain and
the athlete's current form.

A gap exists for an accessible prediction method that accounts for route difficulty and individual
fitness without requiring specialized measurements. Machine learning offers a potential solution:
by learning from an athlete's historical performance data, a model can implicitly capture the
relationship between route characteristics, fitness state, and resulting duration.

\subsection{Objectives}

The main objective of this work is to develop a machine learning model capable of predicting
cycling duration from route topology features and athlete fitness state.

The specific objectives are:

\begin{enumerate}
    \item Combine route topology features with training load metrics as model inputs.
    \item Demonstrate the feasibility of personalized prediction using single-athlete historical
          data in an N-of-1 study design.
    \item Evaluate checkpoint-based progressive predictions that update estimates as the ride
          progresses.
    \item Identify the most predictive features through importance analysis.
\end{enumerate}

\subsection{N-of-1 Study Design Rationale}

This work employs an N-of-1 study design, analyzing 96 rides from a single athlete.
This approach is justified by three factors: (1) \textbf{High inter-individual variability}
in cycling physiology---power profiles, VO$_2$max, lactate thresholds, and anaerobic capacity
vary 2-5x across amateur cyclists. Population models would require extensive physiological
testing (laboratory VO$_2$max, critical power profiling) impractical for the target use case.
(2) \textbf{Personalized prediction goal}---we seek to answer ``How long will this route
take ME?'' rather than ``How long for an average cyclist?'' Personalized models learn
athlete-specific pacing patterns and strengths. (3) \textbf{Data availability}---modern
cyclists generate thousands of GPS activities, providing abundant within-subject data
without requiring sparse multi-athlete cohorts. This design has proven effective in
chronobiology, nutrition, and clinical medicine for personalized prediction tasks with
high inter-subject variability.

\subsection{Paper Structure}

The remainder of this paper is organized as follows. Section 2 reviews related work in cycling
performance prediction, machine learning for sports analytics, and training load modeling.
Section 3 describes the methodology, including data collection, feature engineering, and model
architecture. Section 4 presents the experimental evaluation and results. Section 5 demonstrates
practical application through a case study. Section 6 discusses the findings, limitations, and
directions for future work.

\section{Related Work}

\subsection{Cycling Performance Prediction}

The prediction of cycling performance has traditionally relied on physics-based models.
Di Prampero et al.~\cite{diprampero1979equation} established the foundational equation of
motion for cyclists, relating power output to resisting forces: aerodynamic drag, rolling
resistance, gravitational force on gradients, and drivetrain losses. Martin et al.~\cite{martin1998validation}
validated this mathematical model for road cycling power, demonstrating accurate prediction
of power requirements under controlled conditions. This equation forms the basis of commercial
tools that predict race times and optimal pacing strategies.

Physics-based approaches achieve high accuracy when properly parameterized. However, they require
inputs that are difficult to obtain in practice. The aerodynamic drag coefficient (CdA) depends on
rider position, equipment, and clothing; measuring it accurately requires wind tunnel testing or
specialized field protocols. Rolling resistance varies with tire pressure, surface conditions, and
temperature. Wind speed and direction along a route are rarely available with sufficient spatial
and temporal resolution.

Critical Power (CP) models offer an alternative framework, characterizing an athlete's capacity
through two parameters: CP (the power sustainable indefinitely) and W' (the finite work capacity
above CP). These models predict time to exhaustion at given intensities but do not directly address
the route-specific duration prediction problem, as they assume constant power output.

\subsection{Machine Learning in Sports Analytics}

Machine learning has been applied extensively to sports prediction problems~\cite{chen2020machine}.
In cycling, researchers have used gradient boosting methods to predict race outcomes, leveraging
historical results, course profiles, and athlete rankings. Deep learning approaches, including
LSTM networks, have modeled sequential performance data to capture temporal patterns in athlete form.
Jobson et al.~\cite{jobson2009prediction} demonstrated the utility of cycling training data analysis
for performance modeling, while Menaspà et al.~\cite{menaspa2017cycling} characterized the physical
demands of professional cycling using power meter data.

A notable distinction exists between predicting rankings and predicting actual times. Most published
work focuses on classification or ranking tasks: determining which athlete will finish first or
estimating finishing position. Direct time prediction receives less attention, likely due to the
additional complexity of modeling absolute performance rather than relative performance.

For running, the TRAP framework demonstrated checkpoint-based progressive prediction, where estimates
are refined as an athlete passes intermediate timing points. This approach has not been systematically
applied to cycling, where checkpoint infrastructure is less common and routes are more variable.

\subsection{Training Load and Fitness Modeling}

The relationship between training and performance has been modeled through fitness-fatigue frameworks.
The impulse-response model represents an athlete's performance potential as the difference between
a positive fitness component and a negative fatigue component, both responding to training stimuli
with different time constants.

Training Stress Score quantifies the physiological load of individual workouts based on
intensity and duration relative to threshold power~\cite{coggan2019tss}. Chronic Training Load
(CTL) represents the exponentially weighted average of Training Stress Score over approximately 42 days, serving
as a proxy for fitness. Acute Training Load (ATL) uses a shorter time constant (approximately
7 days) to capture recent fatigue. Training Stress Balance (TSB), calculated as CTL minus ATL,
indicates an athlete's freshness and theoretical readiness to perform.

These metrics are widely used in training prescription and have been incorporated into performance
prediction models. However, their integration with route-specific features for duration prediction
remains underexplored.

\subsection{Research Gap and Contribution}

Existing approaches to cycling duration prediction either require extensive parameterization
(physics-based models) or focus on ranking rather than absolute time prediction (machine learning
approaches). The combination of route topology features with training load metrics for direct
duration prediction represents an unexplored area.

This work addresses the gap by proposing an accessible machine learning approach that:
\begin{itemize}
    \item Uses readily available data: route files (GPX) and training history from standard
          cycling computers and platforms.
    \item Learns athlete-specific performance patterns without requiring aerodynamic or
          physiological testing.
    \item Incorporates fitness state through established training load metrics.
    \item Supports checkpoint-based progressive prediction for real-time updates during rides.
\end{itemize}

The N-of-1 study design acknowledges that individual variation in cycling performance makes
personalized models more practical than population-level approaches for time prediction.

\section{Methodology}

\subsection{Data Sources}

The data for this study come from a single amateur cyclist's training records, collected over
multiple years through standard cycling equipment. All data are sourced from Intervals.icu, a
training analysis platform that aggregates data from cycling computers and power meters.

\subsubsection{Data Platform}

Raw data are processed through a medallion data architecture. The ingestion layer
(bronze) retrieves data from the Intervals.icu API; a cleaning pipeline (silver)
applies type coercion, schema validation, and unit standardization. The platform
runs daily, providing clean, consistently formatted data for analysis.

Data cleaning filters activities to ensure quality and relevance: (1) outdoor rides
only (indoor trainer sessions with zero GPS movement excluded), (2) minimum duration
of 30 minutes to exclude short commutes and equipment tests, (3) complete GPS and
altitude streams (activities with $>$10\% missing data points excluded), and (4) rides
with sufficient effort (excluding recovery spins below 0.5 IF). These criteria
ensure the model learns from representative training and racing efforts rather than
casual rides. The final dataset contains 96 activities spanning 31 months (March
2023--October 2025).

\subsubsection{Activities}

The activities dataset contains metadata for each recorded ride: start time, elapsed time,
moving time (time spent in motion, excluding stops), distance, and activity type. Rides are
distinguished from other activity types (runs, swims, strength training) through the type
field. Each activity has a unique identifier that links to the corresponding sensor streams.
Moving time serves as the prediction target throughout this work.

\subsubsection{Wellness and Training Load}

Daily wellness records capture physiological state: body weight, CTL, ATL, and TSB. These are computed
using established formulas based on historical power data and reflect the athlete's
state at the start of each day.

\subsubsection{Activity Streams}

Each activity includes second-by-second sensor data: GPS coordinates, altitude, speed,
power, heart rate, and cadence. Route topology features are derived from altitude and
GPS streams; performance metrics from power and speed data.

\subsection{Feature Engineering}

Table~\ref{tab:features} summarizes all 79 engineered features with their formulas.

\begin{table}[H]
\centering
\small
\resizebox{\textwidth}{!}{%
\begin{tabular}{@{}l l l p{6cm}@{}}
\hline
\textbf{Category} & \textbf{Feature} & \textbf{Formula} & \textbf{Description} \\
\hline
\multicolumn{4}{@{}l}{\textit{Basic Metrics (7)}} \\
& total\_distance & --- & Route distance (km) \\
& total\_ascent & $\sum \max(0, \Delta h)$ & Cumulative elevation gain (m) \\
& total\_descent & $\sum \max(0, -\Delta h)$ & Cumulative elevation loss (m) \\
& elevation\_min/max/avg & --- & Elevation statistics (m) \\
& elevation\_gain\_per\_km & $G / d$ & Ascent normalized by distance (m/km) \\
\hline
\multicolumn{4}{@{}l}{\textit{Gradient Variability (3)}} \\
& punchiness\_score & $\sigma(|\Delta g|)$ & Std of gradient changes \\
& gradient\_std & $\sigma(g)$ & Standard deviation of gradient \\
& gradient\_cv & $\sigma(g) / |\bar{g}|$ & Coefficient of variation \\
\hline
\multicolumn{4}{@{}l}{\textit{Climb Detection (9)}} \\
& num\_climbs & count & Detected climbs (ClimbPro algorithm) \\
& num\_hc/cat1/.../cat4 & --- & Climbs by category (HC $\geq$80k, ..., Cat4 $\geq$8k) \\
& total\_climb\_score & $\sum d_i \times g_i$ & Sum of Garmin scores (Eq.~\ref{eq:climb_score}) \\
& max\_climb\_score & $\max(d_i \times g_i)$ & Hardest individual climb \\
& climb\_density & $n_{\text{climbs}} / d$ & Climbs per km \\
\hline
\multicolumn{4}{@{}l}{\textit{Climb Characteristics (4)}} \\
& avg\_climb\_gradient & $\sum(g_i d_i) / \sum d_i$ & Distance-weighted avg gradient (\%) \\
& avg/max/total\_climb\_length & --- & Climb distance metrics (m) \\
\hline
\multicolumn{4}{@{}l}{\textit{Gradient Distribution (5)}} \\
& pct\_slope\_* & $d_{\text{bucket}} / d$ & \% route in 6 gradient buckets \\
& pct\_above\_5/8/10\% & $d_{>x} / d$ & \% route above threshold \\
\hline
\multicolumn{4}{@{}l}{\textit{Technical Features (6)}} \\
& num\_sharp\_turns & count$(\Delta\theta > 45^\circ)$ & Sharp turns detected \\
& turn\_density & $n_{\text{turns}} / d$ & Sharp turns per km \\
& recovery\_distance & $\sum d_i \cdot \mathbf{1}_{|g|<2\%}$ & Flat distance after climbs (m) \\
& technical\_descent & $\sum d_i \cdot \mathbf{1}_{g<-5\%, \Delta\theta>45^\circ}$ & Steep descent + turn combos (m) \\
& max\_sustained\_gradient & $\max(\bar{g}_{500\text{m}})$ & Max 500m rolling avg gradient (\%) \\
& longest\_climb\_distance & --- & Longest detected climb (m) \\
\hline
\multicolumn{4}{@{}l}{\textit{Athlete Fitness (4)}} \\
& ctl & $\text{EMA}_{42}(\text{TSS})$ & Chronic Training Load (fitness) \\
& atl & $\text{EMA}_{7}(\text{TSS})$ & Acute Training Load (fatigue) \\
& tsb & $\text{CTL} - \text{ATL}$ & Training Stress Balance (form) \\
& ramp\_rate & $\Delta \text{CTL} / \Delta t$ & Fitness rate of change \\
\hline
\multicolumn{4}{@{}l}{\textit{Rolling Zone Hours (48)}} \\
& rolling\_*d\_power\_z*\_hours & $H_{w,z}$ (Eq.~\ref{eq:zones}) & 7 power zones $\times$ 4 windows \\
& rolling\_*d\_hr\_z*\_hours & $H_{w,z}$ (Eq.~\ref{eq:zones}) & 5 heart rate zones $\times$ 4 windows \\
\hline
\multicolumn{3}{@{}l}{\textbf{Total}} & \textbf{79 features} \\
\hline
\end{tabular}%
}
\caption{Complete feature set: route topology (27) and athlete state (52). Notation: $d$ = distance, $g$ = gradient, $h$ = altitude, $\theta$ = bearing, $\Delta$ = difference, $\sigma$ = std dev, $\mathbf{1}$ = indicator.}
\label{tab:features}
\end{table}

\begin{figure}[H]
\centering
\includegraphics[width=\textwidth]{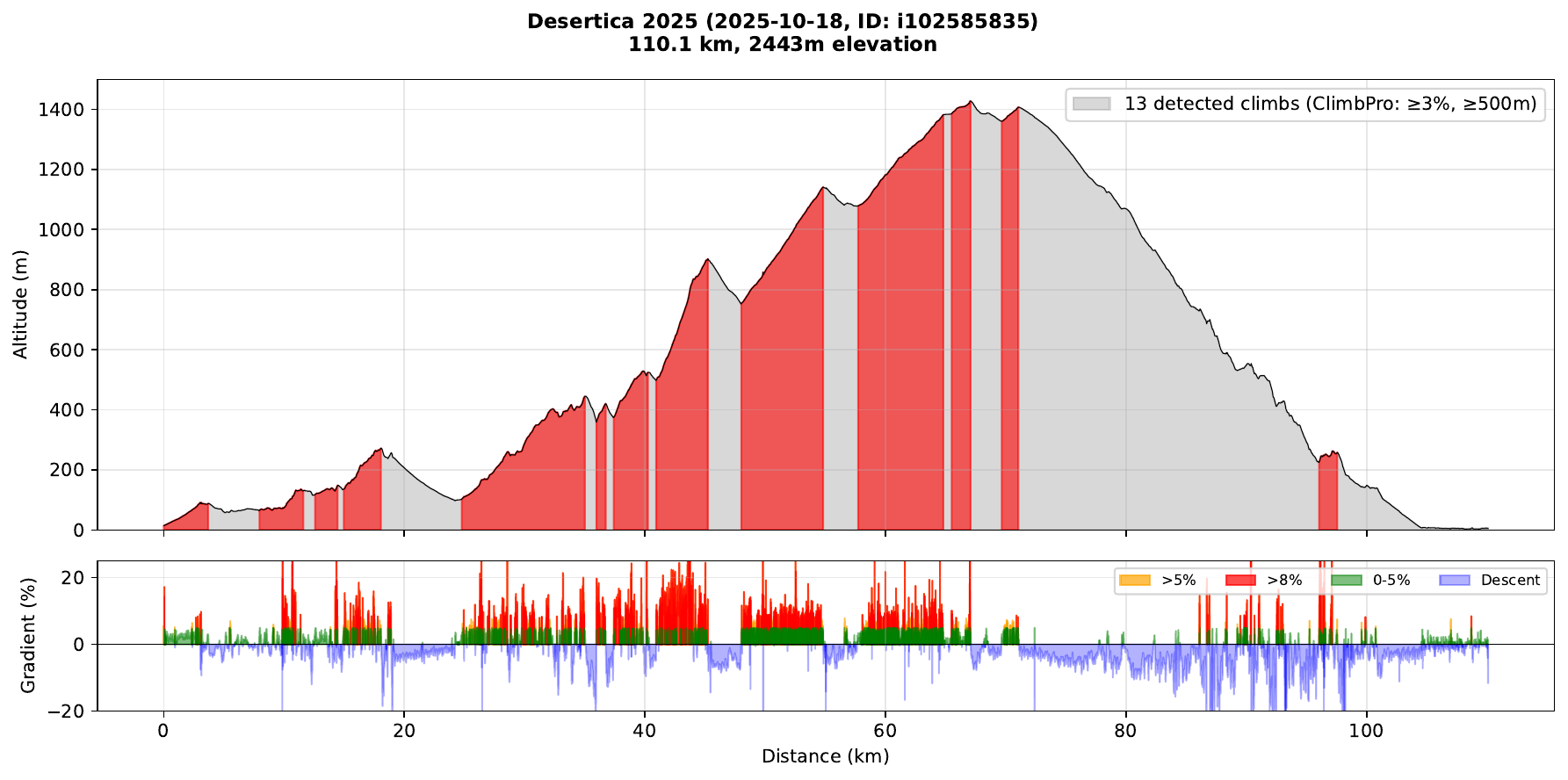}
\caption{Elevation profile of La Desértica MTB race (110 km, 2443m ascent). Top: altitude
with 13 detected climbs (ClimbPro: $\geq$3\%, $\geq$500m) highlighted in red.
Bottom: gradient distribution showing sections above 5\% (orange) and 8\% (red).}
\label{fig:elevation}
\end{figure}

The model inputs are organized into two categories: route topology features that
characterize the physical demands of a course, and athlete state features that capture
training history. Table~\ref{tab:features} summarizes the feature groups.

\subsubsection{Route Topology Features}

Route topology features quantify difficulty from GPS and altitude data. Beyond basic
metrics (distance, elevation), we derive features capturing terrain nuances.
Figure~\ref{fig:elevation} illustrates these features on an example route.

\paragraph{Climb Detection}
Following Garmin ClimbPro's algorithm, climbs are continuous segments where
gradient exceeds 3\% for at least 500m. Each is scored as:
\begin{equation}
S_\mathrm{climb} = d \times g
\label{eq:climb_score}
\end{equation}
where $d$ is segment length in meters and $g$ is average gradient in percentage
points (e.g., 8 for 8\%), yielding an HC score of 80,000 for a 10km climb at 8\%.
Scores above 80,000
are \textit{Hors Catégorie} (HC); 64k, 32k, 16k, 8k for Cat 1--4; $>$1,500 for
uncategorized climbs. We additionally compute Tour de France scores
($g^2 \times d$) as a separate feature. From detected climbs we derive: count by
category (HC, Cat 1--4), total and maximum scores, average gradient weighted by
distance, climb density (climbs per km), and length metrics (average, maximum,
total climb distance). Short climbs (1--5 min) tax anaerobic capacity; sustained
climbs ($>$10 min) depend on aerobic power~\cite{leo2022power}.

\paragraph{Gradient Variability}

\begin{figure}[H]
\centering
\includegraphics[width=\textwidth]{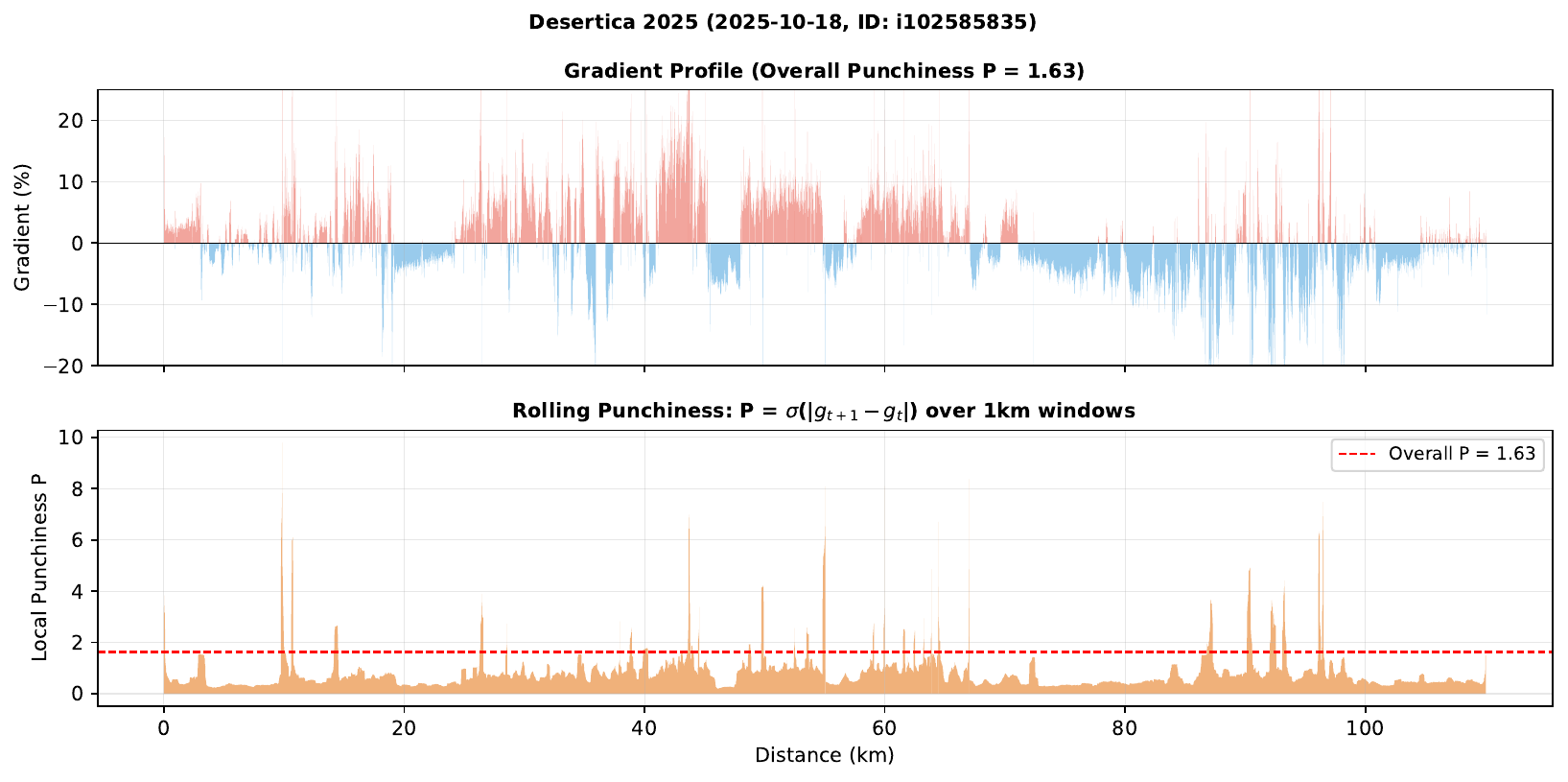}
\caption{Punchiness score analysis. Top: gradient profile with overall route punchiness
$P$ (standard deviation of gradient changes). Bottom: rolling punchiness over 1km windows
for visualization, identifying sections with highest gradient irregularity. The model uses
the global $P$ as a single feature; the rolling view illustrates where variability concentrates.}
\label{fig:punchiness}
\end{figure}

Variable terrain causes 26\% greater fatigue than constant grades at equal average
power~\cite{leo2022durability}. This occurs because gradient variability forces
repeated anaerobic efforts during accelerations on steep sections, prevents
steady-state metabolic adaptation, and requires constant power adjustments that
increase neuromuscular fatigue. We compute a \textit{punchiness score}:
\begin{equation}
P = \text{std}(\Delta g) \quad \text{where} \quad \Delta g_t = |g_{t+1} - g_t|
\end{equation}
The punchiness score $P$ is the standard deviation of absolute gradient changes
across the entire route, yielding a single scalar that captures terrain irregularity.
Unlike gradient standard deviation (which measures overall spread), punchiness
captures the \textit{frequency and magnitude of changes}, making it more sensitive
to rolling terrain that appears flat by traditional metrics. Higher values indicate
rolling terrain with repeated accelerations.
Figure~\ref{fig:punchiness} visualizes local 1km rolling windows to show where
variability is highest, while the model uses the global standard deviation as a feature.

\begin{figure}[H]
\centering
\includegraphics[width=0.85\textwidth]{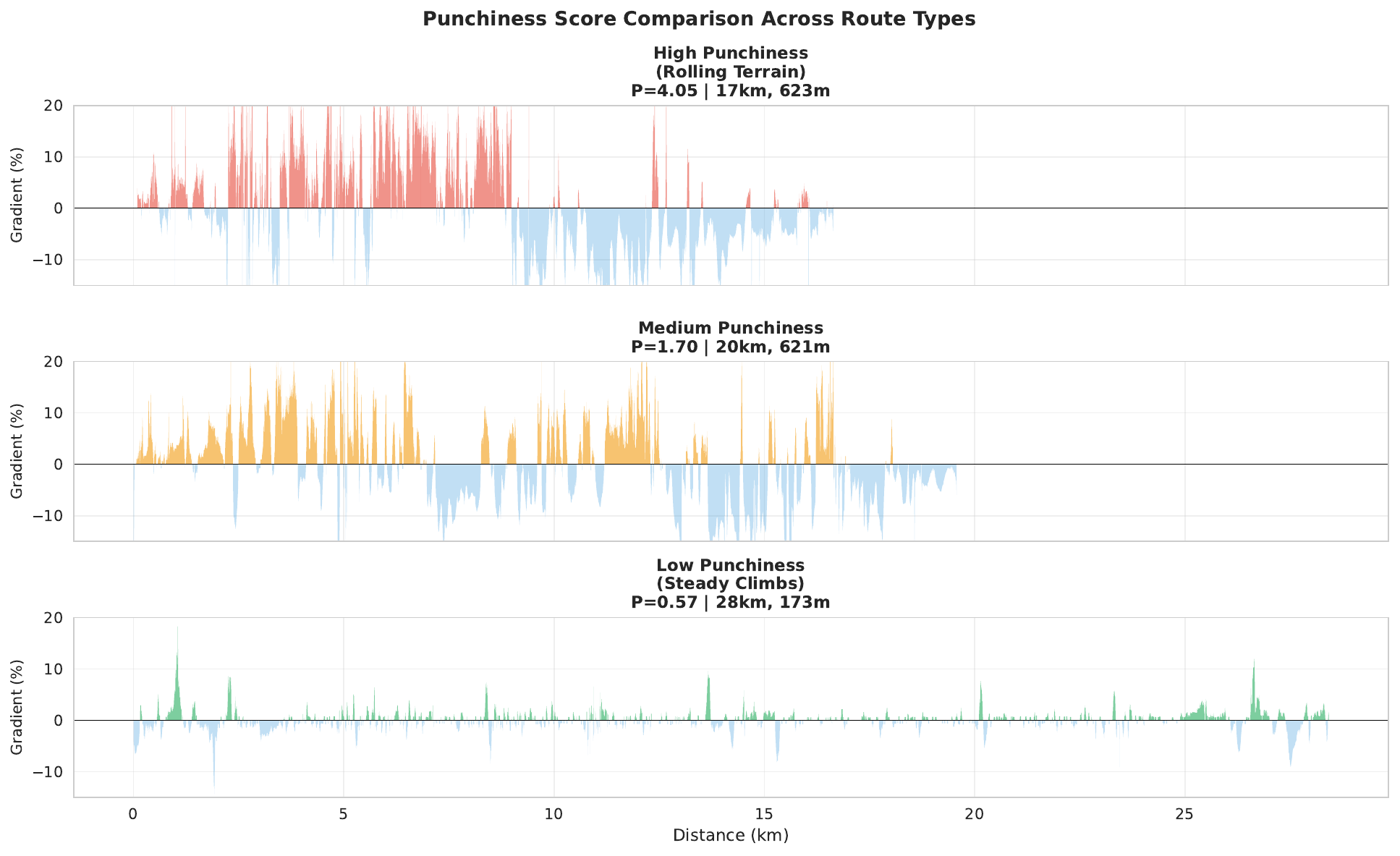}
\caption{Punchiness score comparison across three routes with similar distance and
elevation but different gradient variability. High punchiness (left) indicates frequent
gradient changes requiring repeated power surges; low punchiness (right) shows steady
grades allowing constant pacing. This metric captures difficulty beyond total elevation.}
\label{fig:punchiness_comparison}
\end{figure}

\paragraph{Recovery and Technicality}
\label{sec:recovery}
After significant climbs, flat sections allow W$'$ (anaerobic work capacity)
reconstitution with time constants of 377--580s~\cite{skiba2015intramuscular}.
We compute recovery distance as flat terrain within 500m of climb ends:
\begin{equation}
D_{\text{recovery}} = \sum_{i \in \mathcal{R}} d_i \cdot \mathbf{1}_{|g_i| < 2\%}
\label{eq:recovery}
\end{equation}
where $\mathcal{R}$ is the set of points within 500m after climbs with
$g_{\text{avg}} > 3\%$. Routes with minimal recovery distance accumulate
fatigue faster as W$'$ cannot reconstitute between efforts.

Technical descents are identified where gradient $<-5\%$ coincides with
bearing change $>45^\circ$:
\begin{equation}
D_{\text{tech}} = \sum d_i \cdot \mathbf{1}_{g_i < -5\%} \cdot \mathbf{1}_{\Delta\theta_i > 45^\circ}
\label{eq:technical}
\end{equation}
These sections require sustained attention despite low power output,
limiting recovery~\cite{abbiss2008describing}.

\begin{figure}[H]
\centering
\includegraphics[width=\textwidth]{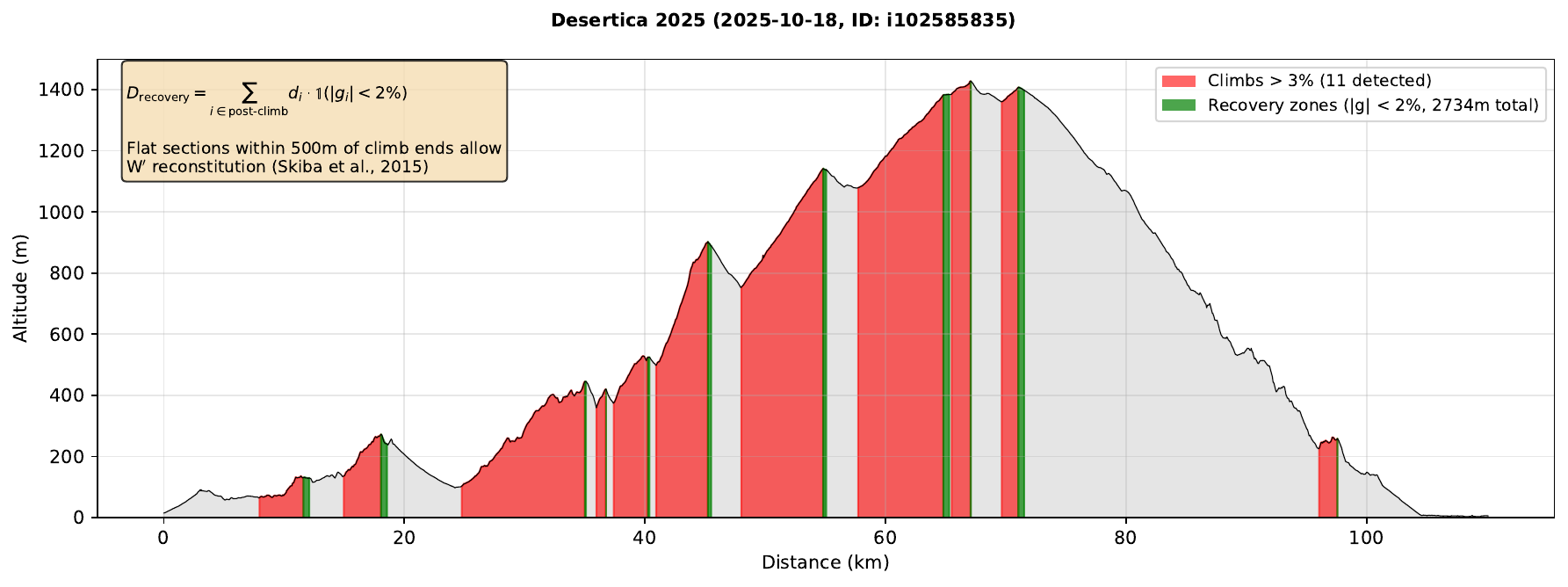}
\caption{Recovery distance analysis. Climbs (red, $g > 3\%$) and recovery zones
(green, $|g| < 2\%$ within 500m of climb ends). Flat sections after climbs allow
W$'$ reconstitution; routes with minimal recovery distance accumulate fatigue faster.}
\label{fig:recovery}
\end{figure}

\begin{figure}[H]
\centering
\includegraphics[width=0.9\textwidth]{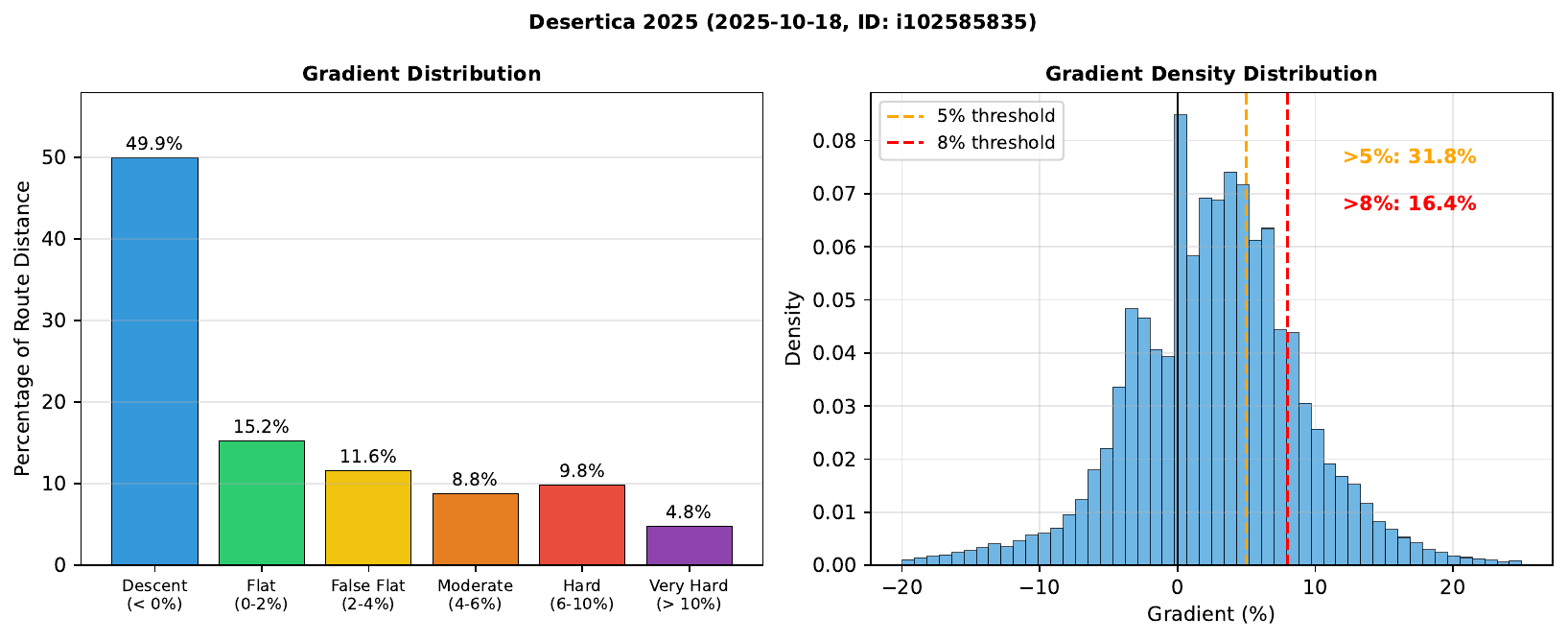}
\caption{Gradient distribution analysis showing three complementary views: histogram
of gradient values, cumulative distance in gradient buckets, and sustained gradient
over rolling 500m windows. Routes with similar total elevation can have vastly
different gradient profiles, affecting physiological demand.}
\label{fig:gradient_dist}
\end{figure}

\paragraph{Gradient Distribution}
Physiological response varies non-linearly with gradient, with steep sections (>6\%)
requiring anaerobic contributions and moderate gradients (2-6\%) sustainable aerobically.
We compute distance percentages in gradient buckets (negative, 0-2\%, 2-4\%, 4-6\%,
6-10\%, 10\%+) and thresholds (\% route above 5\%, 8\%, 10\%). Figure~\ref{fig:gradient_dist}
illustrates gradient analysis on an example route.

\begin{figure}[H]
\centering
\includegraphics[width=0.8\textwidth]{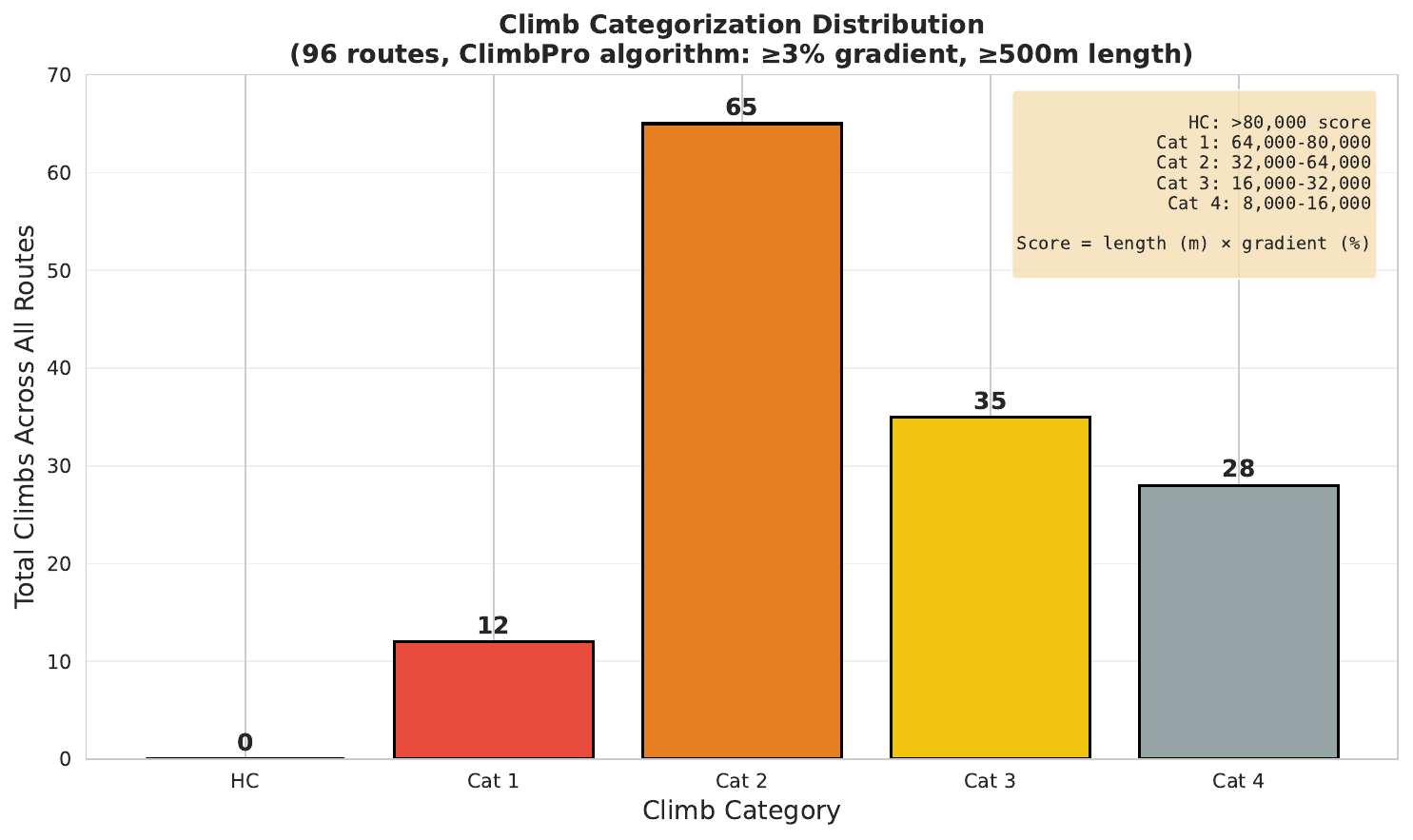}
\caption{Climb categorization for dataset routes using Garmin ClimbPro algorithm.
Most routes contain Cat 3-4 climbs (8-32k score), while challenging routes feature
HC and Cat 1 climbs ($>$64k score). Climb category distribution captures route
difficulty beyond simple elevation gain.}
\label{fig:climb_categories}
\end{figure}

\begin{figure}[H]
\centering
\includegraphics[width=0.95\textwidth]{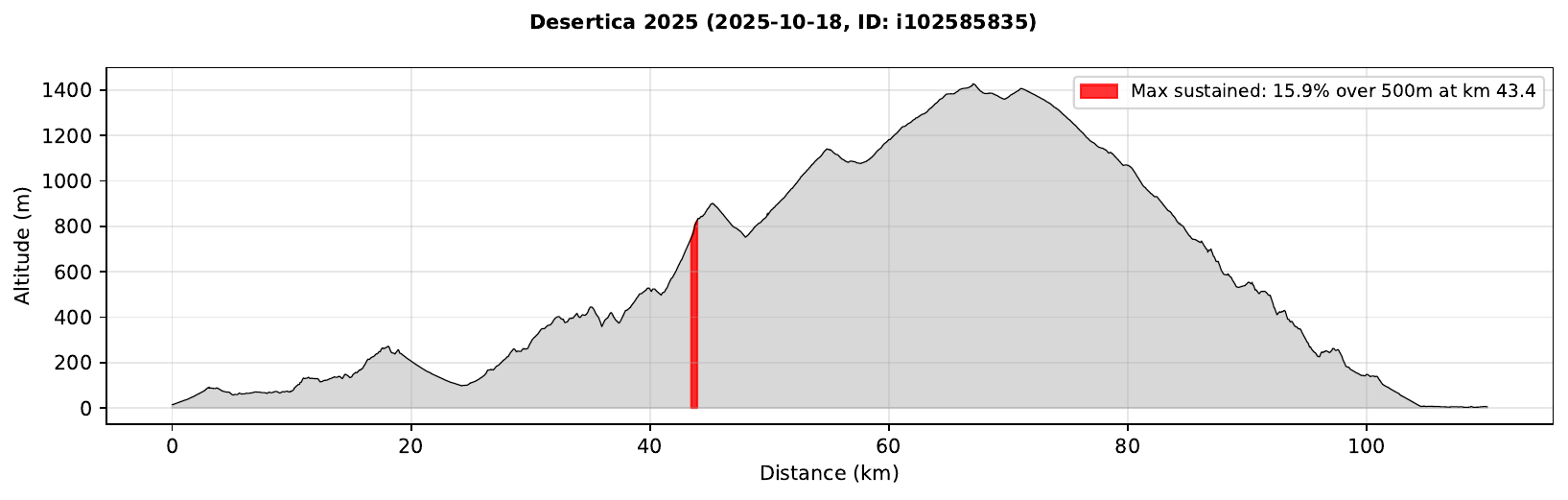}
\caption{Maximum sustained gradient analysis. The 500m section with highest average
gradient (peak physiological demand) is highlighted on the elevation profile, showing
location of hardest sustained effort within the route.}
\label{fig:max_gradient_location}
\end{figure}

\paragraph{Elevation Profile Shape}
We also compute route
distance in six gradient buckets (descending, flat, false flat, moderate, hard, very hard)
and percentage of total ascent in each route third. Back-loaded routes may induce greater
fatigue from glycogen depletion.

\paragraph{Sharp Turns and Turn Density}
\label{sec:turns}
Sharp turns require braking, gear changes, and power surges to accelerate out,
disrupting rhythm and accumulating fatigue~\cite{padilla2000scientific}. We detect
sharp turns where bearing change exceeds $45^\circ$:
\begin{equation}
\Delta\theta_t = \min\bigl(|\theta_{t+1} - \theta_t|,\, 360^\circ - |\theta_{t+1} - \theta_t|\bigr)
\label{eq:bearing}
\end{equation}
where $\theta_t$ is the smoothed bearing (5-point rolling mean to reduce GPS jitter).
A turn is sharp when $\Delta\theta > 45^\circ$. Turn density normalizes by distance:
$T_d = n_{\text{turns}} / d_{\text{km}}$.

\begin{figure}[H]
\centering
\includegraphics[width=0.95\textwidth]{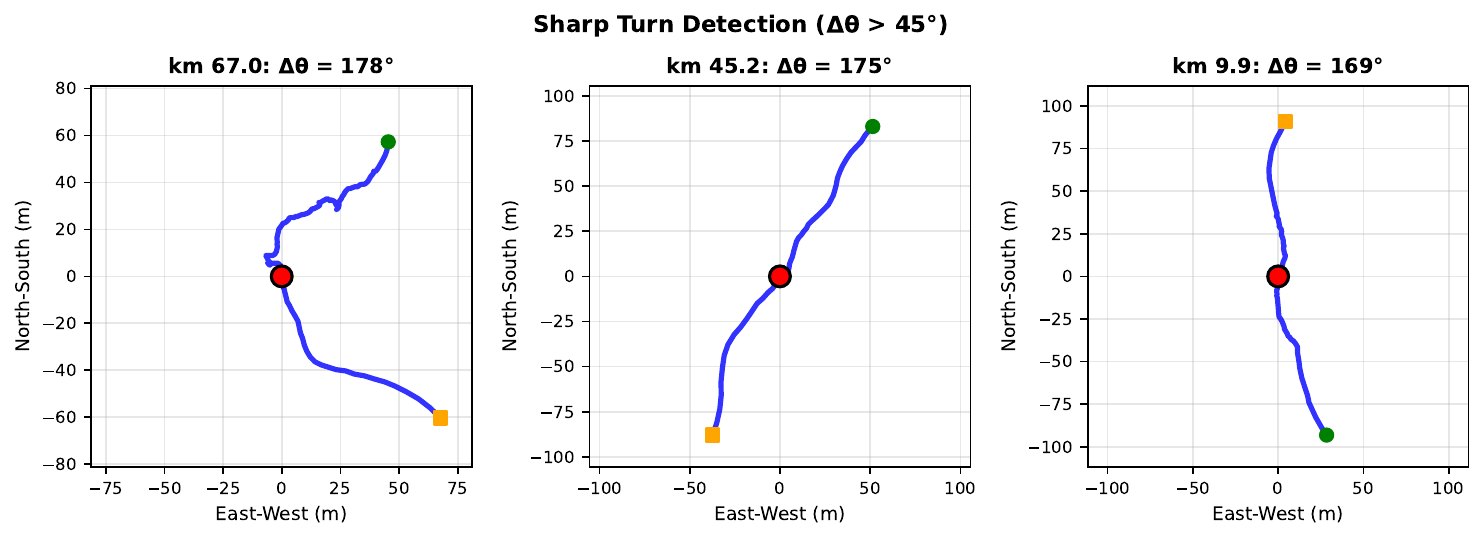}
\caption{Sharp turn detection examples. Each panel shows the route geometry around a
detected turn (red dot at origin) where bearing change exceeds $45^\circ$. Axes show
distance in meters from turn point; green/orange markers indicate travel direction.}
\label{fig:sharp_turns}
\end{figure}

\paragraph{Additional Route Metrics}
Elevation gain per km normalizes climbing by distance, enabling comparison
across routes of different lengths.

\subsubsection{Athlete State Features}

Training intensity distribution predicts performance~\cite{sanders2017methods}.
Following Banister's impulse-response model~\cite{banister1991modeling} and
Seiler's time-in-zone methodology~\cite{seiler2006quantifying}, we compute
rolling zone hours:
\begin{equation}
H_{w,z} = \frac{1}{3600}\sum_{t - w < t' < t} s_{z,t'}
\label{eq:zones}
\end{equation}
where $H_{w,z}$ is hours in zone $z$ over window $w$ days, and $s_{z,t'}$ is
seconds in zone $z$ on day $t'$. Strict inequality prevents data leakage.

\vspace{0.5em}
\noindent\textbf{Windows:} 7d (acute), 14d (short-term), 30d (monthly), 60d (chronic)

\noindent\textbf{Power zones:} Z0--Z6 (7 zones) $\times$ 4 windows = 28 features

\noindent\textbf{Heart rate zones:} Z0--Z4 (5 zones) $\times$ 4 windows = 20 features

\vspace{0.5em}
High hours in upper zones over short windows indicate recent fatigue-inducing
intensity; accumulated endurance zone hours over longer windows reflect aerobic base.

\subsubsection{Training vs. Deployment Feature Sets}

\textbf{Critical distinction:} This work evaluates features for \textit{training} but
deploys models using only \textit{prediction-time available} features.

During model development, we evaluated three feature configurations to determine
what information drives duration prediction:

\begin{enumerate}
\item \textbf{Topology-only} (27 features): Route characteristics from GPS and
altitude streams---distance, elevation, climbs, gradient distribution, punchiness.
All extractable from GPX files before a ride.

\item \textbf{Topology + Fitness} (31 features): Adds athlete state at ride
start---CTL, ATL, TSB, ramp rate. Requires historical training data.

\item \textbf{Topology + Fitness + Zones} (79 features): Adds rolling zone hours
(7d, 14d, 30d, 60d windows across power and heart rate zones). Captures recent training
load distribution.
\end{enumerate}

\textbf{Results:} The best model (Lasso Topology + Fitness) achieves MAE=6.60 min
and R²=0.922 using 31 features (27 topology + 4 fitness). Fitness metrics improve
accuracy by 14\% over topology alone (MAE=7.66 min), demonstrating that physiological
state meaningfully constrains performance even in self-paced efforts.

\textbf{Deployment scenarios:}

\begin{itemize}
\item \textbf{Best accuracy}: Use Topology + Fitness model (31 features) when historical
training data are available. Requires athlete's current CTL, ATL, TSB, and ramp rate from
prior weeks of training.

\item \textbf{Cold start / Universal}: Use Topology-only model (27 features) when predicting
for new athletes without training history, or when analyzing routes without fitness context.
Accepts any GPX file (GPS + elevation) with slightly reduced accuracy (MAE=7.66 vs 6.60 min).

\item \textbf{What-If scenario analysis}: The Topology + Fitness model enables projections
like ``Given this route and my target fitness in 12 months (CTL=65, TSB=+10), what duration
should I expect?'' This is possible because fitness features use historical training load,
not ride performance results.
\end{itemize}

\subsection{Progressive Prediction Framework}

For pre-ride planning, the model predicts duration from the complete route topology.
For race-day applications, we evaluate checkpoint-based progressive predictions where
estimates are updated as the athlete completes portions of the route.

\paragraph{Checkpoint Definition}
Checkpoints are defined at fixed distance percentages (25\%, 50\%, 75\%, 100\%) along
the route. At each checkpoint, topology features are recomputed from the GPS track up
to that point, providing a cumulative route profile up to the current position.

\paragraph{Feature Recalculation}
All topology features (climbs, gradients, punchiness) are recalculated on the truncated
track. For example, at 50\% completion on a 100km route, features are extracted from
the first 50km only. This creates a prediction based on terrain encountered so far,
revealing route difficulty as it unfolds.

\paragraph{Use Case}
Progressive predictions answer: ``Given the terrain I've ridden so far, how much total
time will this route take?'' This differs from real-time remaining time prediction,
instead providing updated full-route duration estimates as more terrain is revealed.
Prediction stability depends on route difficulty distribution---front-loaded climbs yield
stable early estimates; back-loaded routes show greater prediction evolution.

\subsection{Model Selection and Validation}

\subsubsection{Model Candidates}
We evaluate three model families: (1) \textbf{Regularized linear models} (Ridge, Lasso,
ElasticNet) with L1/L2 penalties to prevent overfitting, (2) \textbf{Random Forest} with
Bayesian hyperparameter optimization via Optuna~\cite{akiba2019optuna}, and (3)
\textbf{Baseline models} (mean, median, simple linear regression on distance + elevation)
to establish prediction difficulty.

\subsubsection{Nested Cross-Validation}
With $N=96$ rides and up to 79 features (topology + fitness + zones), overfitting risk
is severe. We employ nested cross-validation to obtain unbiased performance estimates:

\paragraph{Outer Loop (Evaluation)} 5-fold stratified CV by distance quintiles ensures
test folds remain isolated, never used for hyperparameter selection. Stratification by
distance prevents imbalanced folds (long rides concentrated in one fold).

\paragraph{Inner Loop (Hyperparameter Tuning)} For each outer training fold:
\begin{itemize}
\item Linear models: 3-fold CV with automatic alpha selection (\texttt{RidgeCV}, \texttt{LassoCV})
\item Random Forest: 3-fold CV + Optuna Bayesian optimization (30 trials per fold)
\end{itemize}

\paragraph{Leakage Prevention}
\begin{enumerate}
\item Feature imputation (median strategy) fitted on training fold only
\item Standardization (z-score) fitted on training fold only
\item Rolling zone features use strict temporal inequality (Equation~\ref{eq:zones})
\end{enumerate}

\subsubsection{Evaluation Metrics}
Models are compared by: (1) \textbf{MAE} (Mean Absolute Error in minutes), interpretable
for athletes, (2) \textbf{R²}, proportion of variance explained, (3) \textbf{Train-Test Gap},
difference between train and test MAE detecting overfitting (gap $< -2$ min indicates
severe overfitting), and (4) \textbf{CV Stability}, coefficient of variation across folds
indicating robustness.

\section{Model Evaluation}

\subsection{Dataset}

The dataset comprises 96 outdoor cycling activities recorded between March 2023 and
October 2025 from a single amateur cyclist (male, Functional Threshold Power 232W). Activities were classified
as free rides (N=92) or races (N=4) using duration deviation from expected time.
Table~\ref{tab:dataset_stats} summarizes route and performance characteristics.

\begin{table}[H]
\centering
\small
\begin{tabular}{@{}l r r@{}}
\hline
\textbf{Metric} & \textbf{Mean $\pm$ SD} & \textbf{Range} \\
\hline
Distance (km) & 24.4 $\pm$ 13.1 & [10.2, 110.1] \\
Elevation (m) & 520 $\pm$ 339 & [100, 2,575] \\
Moving Time (min) & 114.1 $\pm$ 57.5 & [55.9, 546.9] \\
Avg Gradient (\%) & 2.1 $\pm$ 0.8 & [0.5, 4.7] \\
Climbs per Ride & 3.2 $\pm$ 2.8 & [0, 13] \\
\hline
\end{tabular}
\caption{Dataset statistics for 96 cycling activities.}
\label{tab:dataset_stats}
\end{table}

\subsubsection{Fitness State Distribution}
\begin{figure}[H]
\centering
\includegraphics[width=0.9\textwidth]{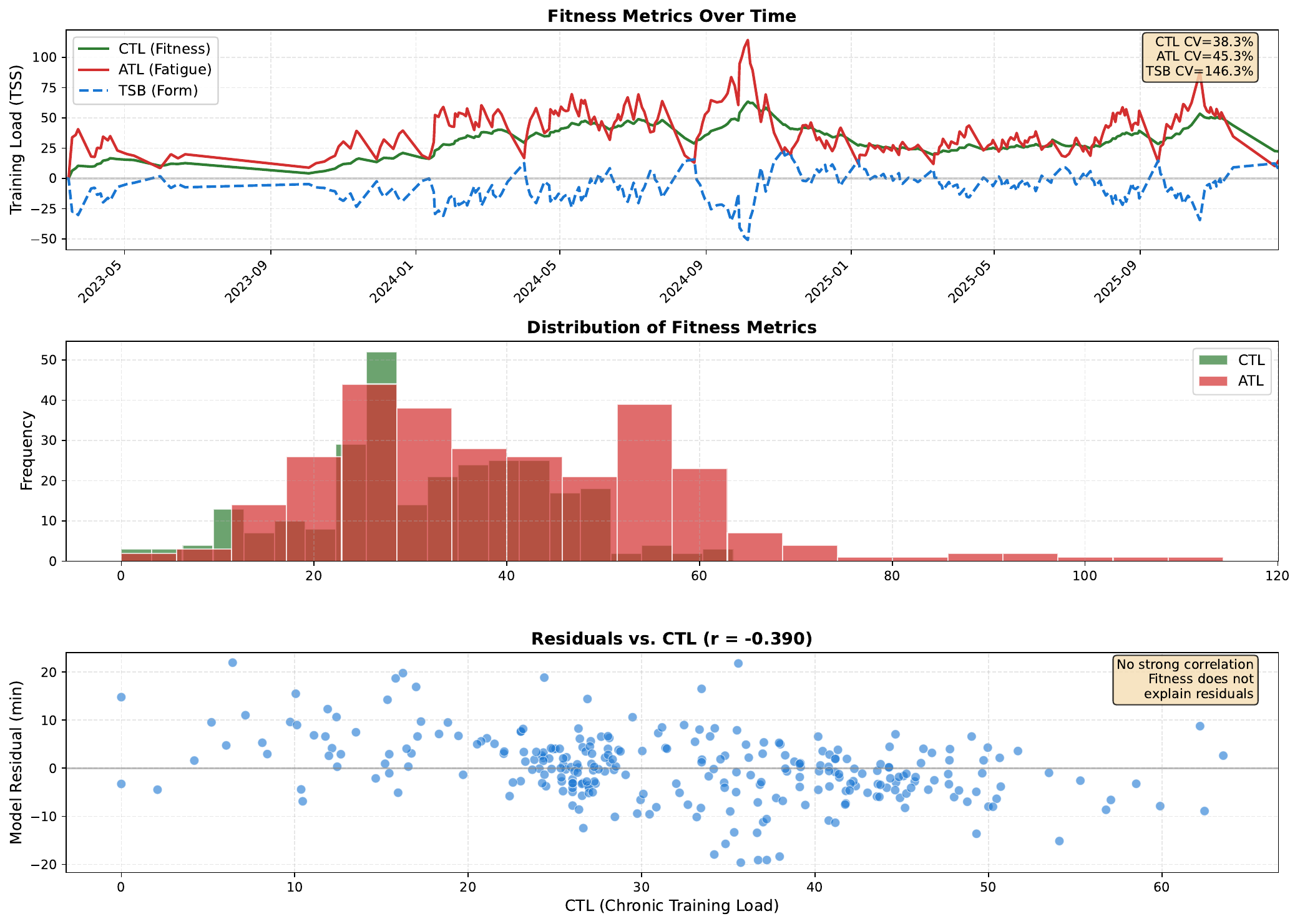}
\caption{Fitness metric temporal analysis. Top: CTL/ATL/TSB time series showing moderate
variation (CV $\approx$ 54\%) over 31-month period. Middle: Distribution histograms.
Bottom: Model residuals vs. CTL showing weak correlation (r=$-$0.37), suggesting fitness
metrics do not strongly explain prediction errors.}
\label{fig:fitness_temporal}
\end{figure}

Training load metrics exhibit moderate temporal variation (Figure~\ref{fig:fitness_temporal}):
CTL ranges from 0 (training restart after break) to 63.5 (peak fitness), with
CV=54.3\%. TSB ranges from $-58$ (deeply fatigued) to $+15.8$ (well-rested), with
CV=89.5\%. The presence of zero values in early 2023 indicates return from training
break, while sustained CTL $>30$ in late 2023-2024 reflects consistent training volume.

\subsubsection{Dataset Diversity}
The dataset spans diverse route types with different characteristics: short punchy climbs, long steady
grades, and mixed terrain. Routes with similar distances can have vastly different
duration and difficulty due to gradient distribution, climb categories, and punchiness scores.

\subsubsection{Feature Engineering}
\begin{figure}[H]
\centering
\includegraphics[width=0.75\textwidth]{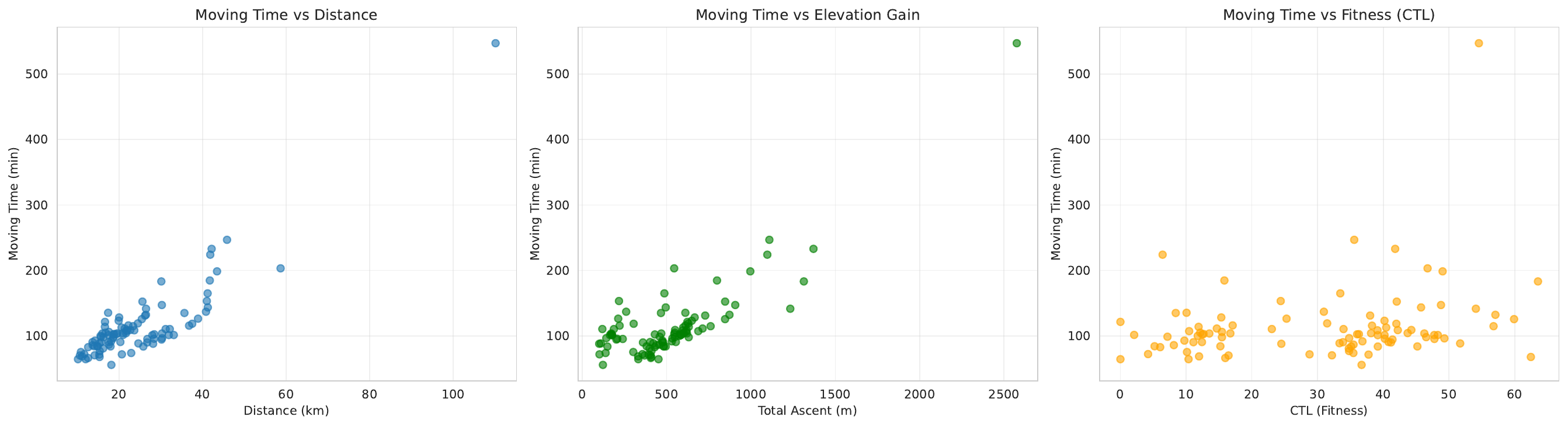}
\caption{Dataset scatter plot showing relationship between distance, elevation gain,
and moving time. Point size indicates number of climbs. Strong correlation visible
between distance/elevation and duration (R²=0.908 for simple linear model), but
significant variance remains unexplained without terrain complexity features.}
\label{fig:duration_scatter}
\end{figure}

From the 96 activities, we extracted 79 features: 27 route topology metrics
(Table~\ref{tab:features}), 4 fitness metrics, and 48 rolling zone hours. Figure~\ref{fig:duration_scatter}
visualizes the relationship between distance, elevation, and duration.

\begin{figure}[H]
\centering
\includegraphics[width=0.9\textwidth]{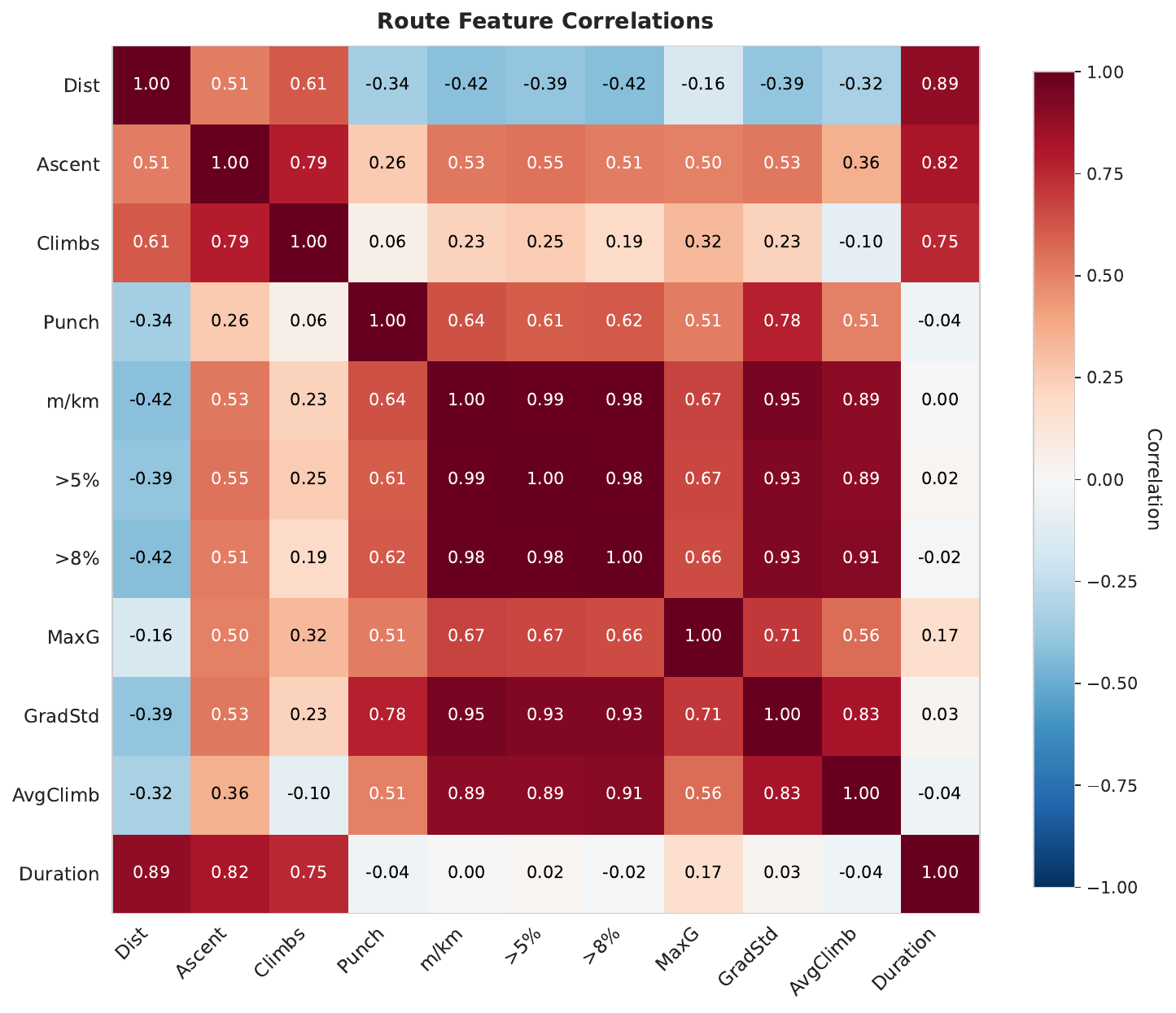}
\caption{Feature correlation matrix for top topology features. Distance and total
ascent show strong correlation (r=0.82), while punchiness and climb density capture
orthogonal route characteristics. Low correlation between many features justifies
inclusion of diverse topology metrics.}
\label{fig:feature_corr}
\end{figure}

\subsection{Model Comparison}

Nested cross-validation results reveal that incorporating fitness state is essential for
accurate prediction. The \textbf{Lasso (Topology + Fitness)} model achieves the best
performance with MAE=6.60 minutes and R²=0.922. This represents a significant improvement
over the Topology-only baseline (MAE=7.66 min). Complex non-linear models (Random Forest)
perform poorly (MAE=13.36 min), suffering from severe overfitting due to the small sample
size (N=96), validating the choice of regularized linear models for this domain.

\begin{figure}[H]
\centering
\includegraphics[width=0.85\textwidth]{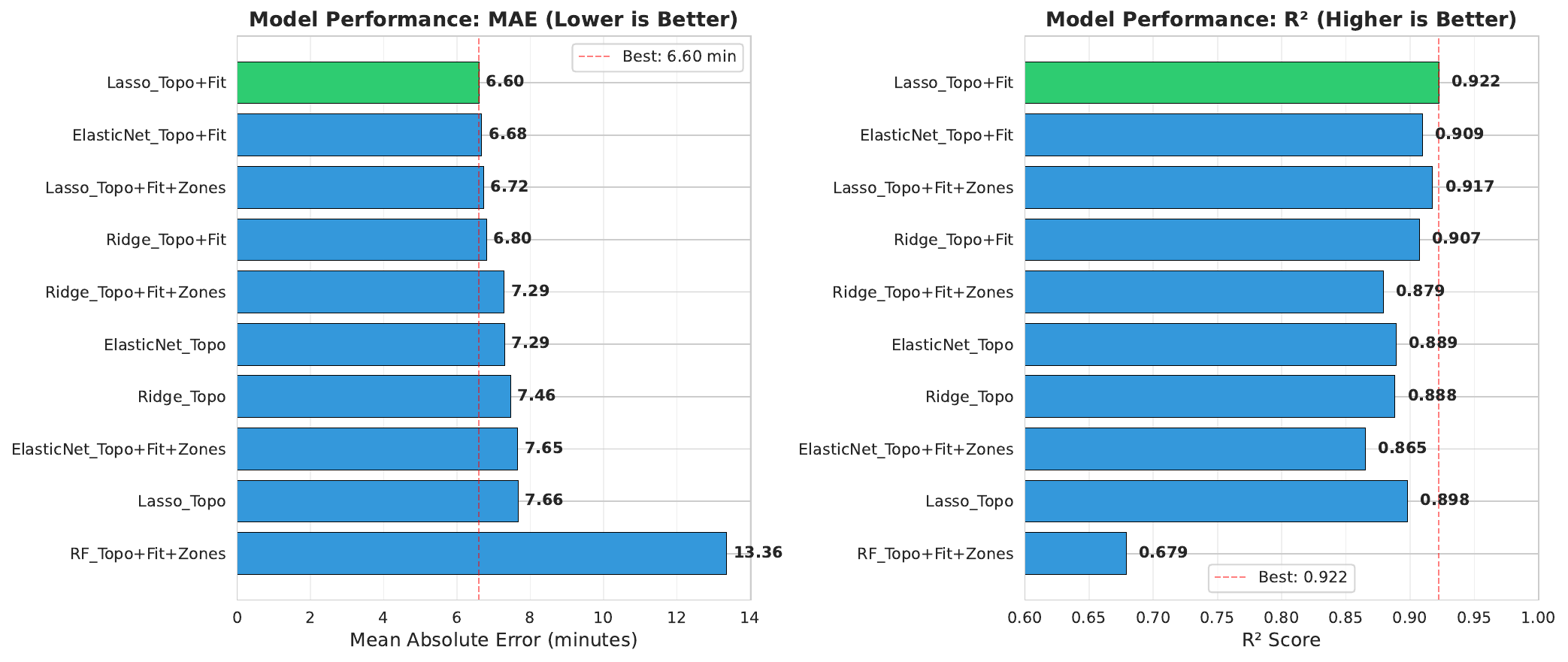}
\caption{Model performance comparison showing MAE and R² across all models. Lasso
(Topology + Fitness) achieves lowest error. Random Forest shows high variance and
severe overfitting despite Optuna hyperparameter tuning. Error bars show standard
deviation across 5 CV folds.}
\label{fig:model_comparison}
\end{figure}

\begin{table}[H]
\centering
\small
\begin{tabular}{@{}l c c c@{}}
\hline
\textbf{Model} & \textbf{MAE (min)} & \textbf{R²} & \textbf{Status} \\
\hline
Baseline (Mean) & 31.82 $\pm$ 8.04 & $-0.43$ $\pm$ 0.71 & --- \\
Linear (Dist+Elev) & 6.91 $\pm$ 1.02 & 0.908 $\pm$ 0.077 & Baseline \\
\hline
Lasso (Topo+Fit) & \textbf{6.60 $\pm$ 1.13} & \textbf{0.922 $\pm$ 0.061} & \textbf{Best} \\
ElasticNet (Topo+Fit) & 6.68 $\pm$ 1.01 & 0.909 $\pm$ 0.093 & Good \\
Lasso (Topo+Fit+Zones) & 6.72 $\pm$ 1.05 & 0.917 $\pm$ 0.069 & Good \\
Ridge (Topo+Fit) & 6.80 $\pm$ 0.85 & 0.907 $\pm$ 0.094 & Good \\
ElasticNet (Topo) & 7.29 $\pm$ 0.51 & 0.889 $\pm$ 0.104 & OK \\
Ridge (Topo) & 7.46 $\pm$ 0.42 & 0.888 $\pm$ 0.101 & OK \\
Lasso (Topo) & 7.66 $\pm$ 1.84 & 0.898 $\pm$ 0.070 & Baseline \\
Random Forest (All) & 13.36 $\pm$ 6.52 & 0.679 $\pm$ 0.211 & Overfit \\
\hline
\end{tabular}
\caption{Model performance via 5-fold nested cross-validation. MAE and R² reported as
mean $\pm$ standard deviation across folds. Adding fitness features improves accuracy
by approximately 14\% over topology alone.}
\label{tab:model_results}
\end{table}

\subsubsection{Fitness Metrics Improve Prediction Accuracy}
Adding Training Load metrics (CTL, ATL, TSB) reduces Mean Absolute Error from 7.66 minutes
(Topology only) to 6.60 minutes (Topology + Fitness)---a 14\% reduction. Note that the
top regularized models (Lasso, ElasticNet, Ridge with fitness features) show overlapping
confidence intervals, so we cannot claim statistical significance for Lasso's superiority;
however, all fitness-augmented models consistently outperform topology-only baselines.
While route topology explains the majority of the variance (R²$\approx$0.88--0.90), the
athlete's fitness state provides additional predictive value. This validates the use of
historical training load (CTL/ATL) as a proxy for current physiological potential.

\subsubsection{Regularization and Feature Engineering}
\begin{figure}[H]
\centering
\includegraphics[width=0.85\textwidth]{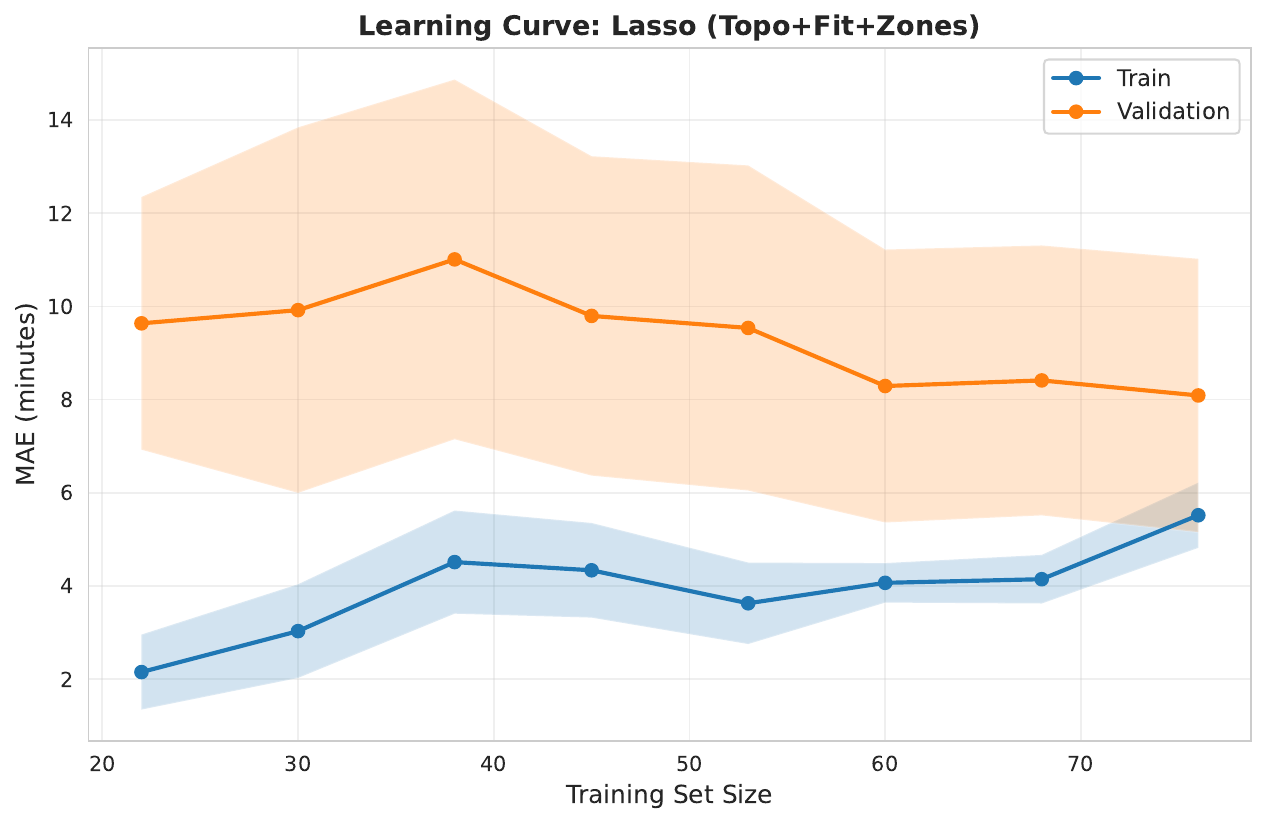}
\caption{Learning curve for Lasso (Topology) model. Validation error (orange) plateaus
at $\sim$60 samples with small train-validation gap, confirming effective regularization
and adequate dataset size. Shaded regions show standard deviation across CV folds.}
\label{fig:learning_curve}
\end{figure}

The disparity between Lasso (MAE=6.60) and Random Forest (MAE=13.36) highlights the
importance of regularization in N-of-1 studies where N is small relative to P features.
By penalizing non-predictive coefficients, Lasso effectively selects a sparse subset of
impactful features, whereas Random Forest memorizes noise in the training folds.
Additionally, the removal of result-dependent features (such as VAM or average speed)
ensures the model is deployment-ready. The reported accuracy relies solely on data
available \textit{prior} to the ride: the GPX file and the athlete's training history.
Figure~\ref{fig:learning_curve} shows that validation error plateaus around 60 samples,
confirming dataset size adequacy for linear models.

\subsubsection{Prediction Accuracy}
\begin{figure}[H]
\centering
\includegraphics[width=0.75\textwidth]{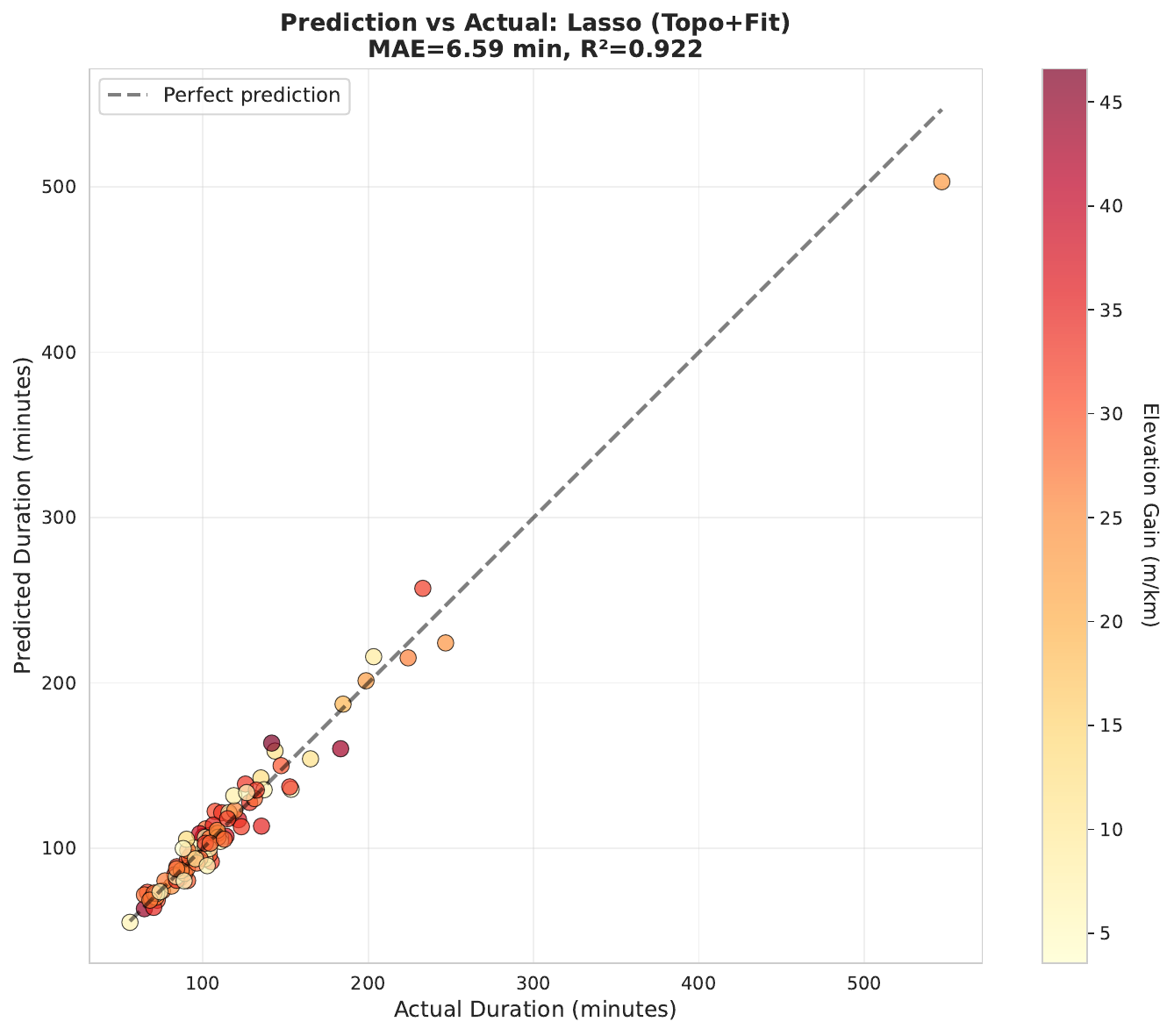}
\caption{Predicted vs. actual moving time for Lasso (Topology + Fitness) model. Points
cluster near diagonal (perfect prediction), with mean absolute error of 6.60 minutes
and R²=0.922 across 96 activities. No systematic bias visible across duration range
(55-547 minutes).}
\label{fig:prediction_actual}
\end{figure}

Figure~\ref{fig:prediction_actual} shows predicted vs. actual durations for the best
model (Lasso Topology + Fitness). Points cluster tightly around the diagonal (perfect
prediction line) with R²=0.922. Residuals show no systematic bias across the duration
range.

\subsection{Error Analysis}

\begin{figure}[H]
\centering
\includegraphics[width=0.85\textwidth]{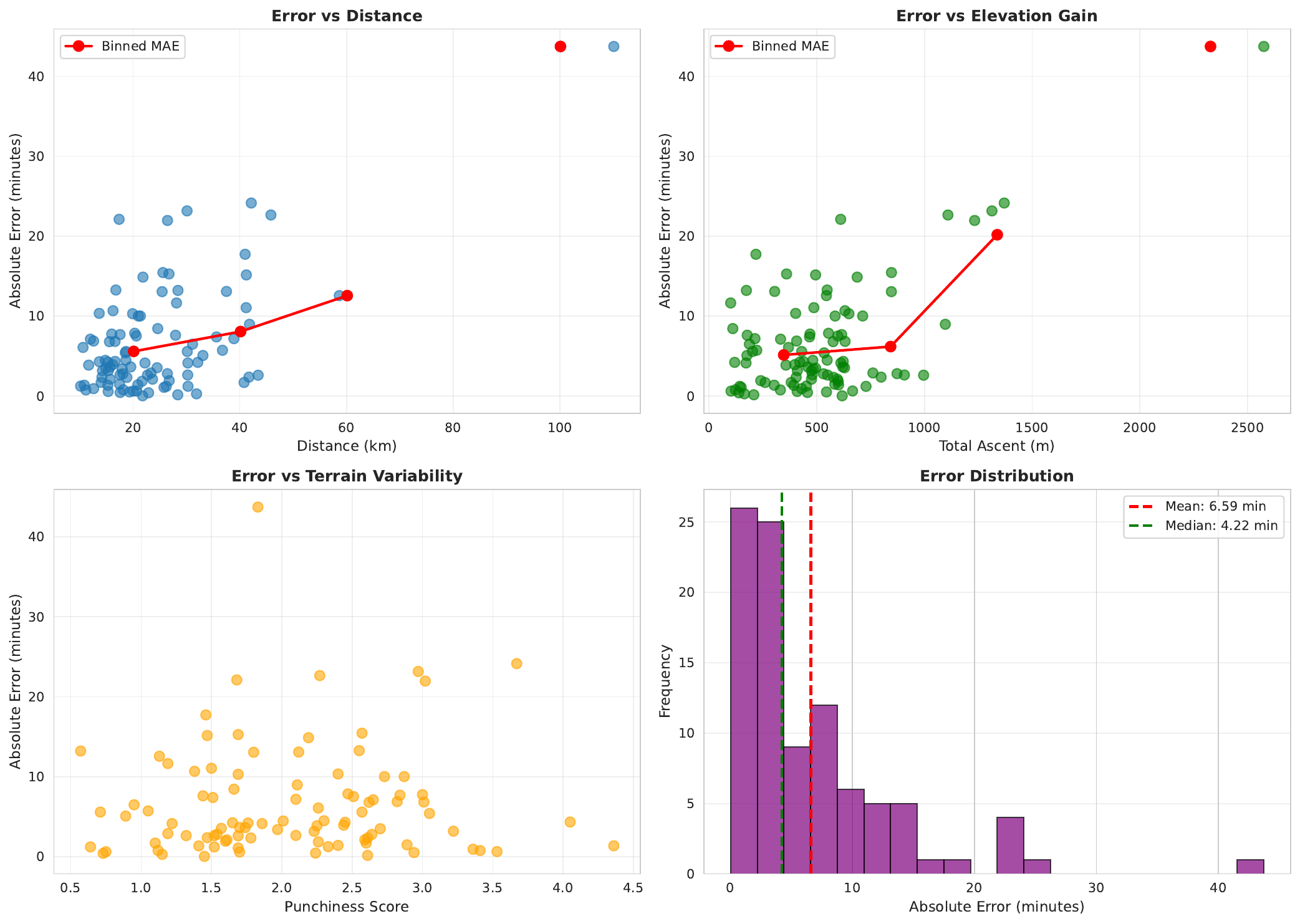}
\caption{Error analysis by route characteristics. Model performs consistently across
difficulty levels with no systematic bias. Larger absolute errors on extreme routes
remain proportionally small (MAPE $<$ 5\% even for longest/steepest routes).}
\label{fig:error_analysis}
\end{figure}

Figure~\ref{fig:error_analysis} breaks down prediction errors by route difficulty.
The model performs consistently across short/long and flat/hilly routes, with no
systematic bias. Larger errors ($>$10 min) occur on extreme routes (very long or
very steep), but remain proportionally small relative to total duration.

\begin{figure}[H]
\centering
\includegraphics[width=0.85\textwidth]{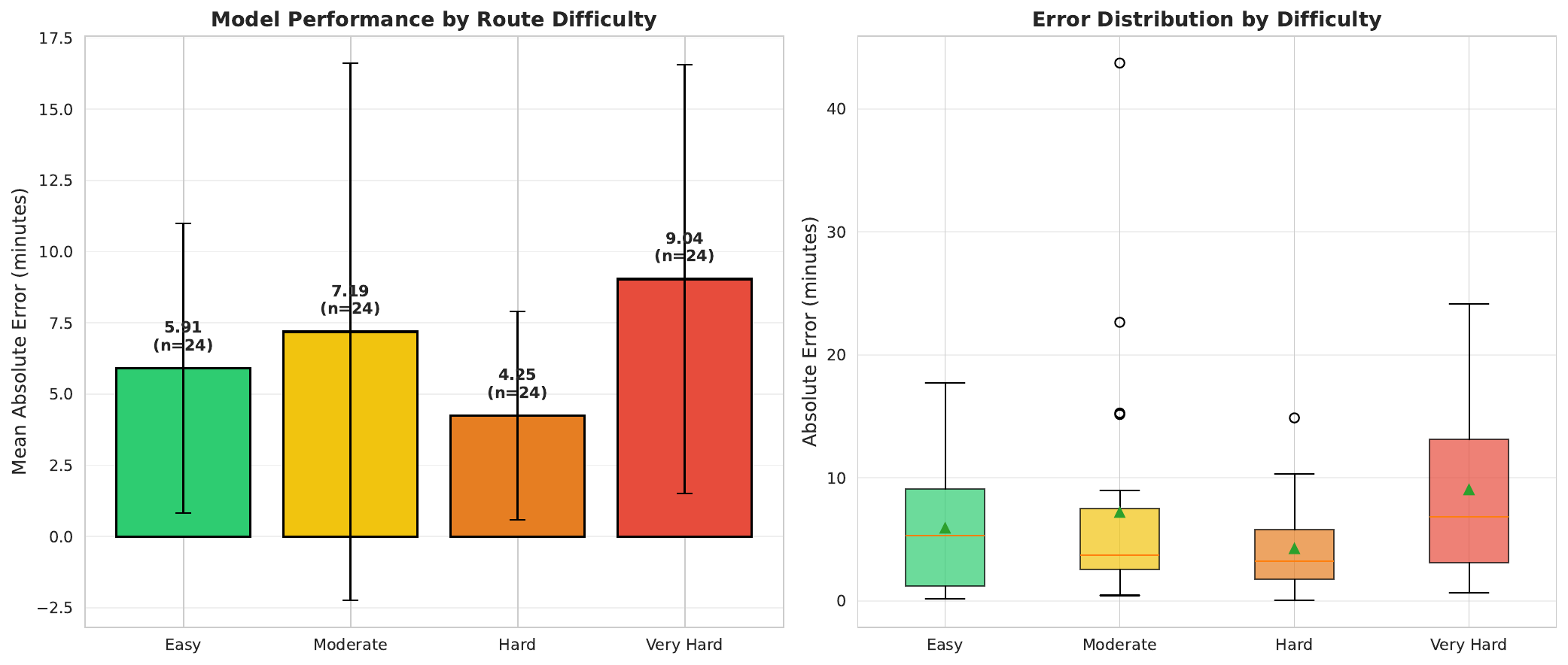}
\caption{Model performance stratified by route difficulty tiers (distance and elevation).
MAE remains stable across easy, moderate, hard, and very hard routes, demonstrating
robustness to varying route characteristics.}
\label{fig:performance_difficulty}
\end{figure}

\subsection{Feature Importance}

To understand which route characteristics drive predictions, we analyze feature importance
using SHAP (SHapley Additive exPlanations)~\cite{lundberg2017unified} values on the
best model (Lasso with topology features).

\subsubsection{Global Importance}
\begin{figure}[H]
\centering
\includegraphics[width=0.85\textwidth]{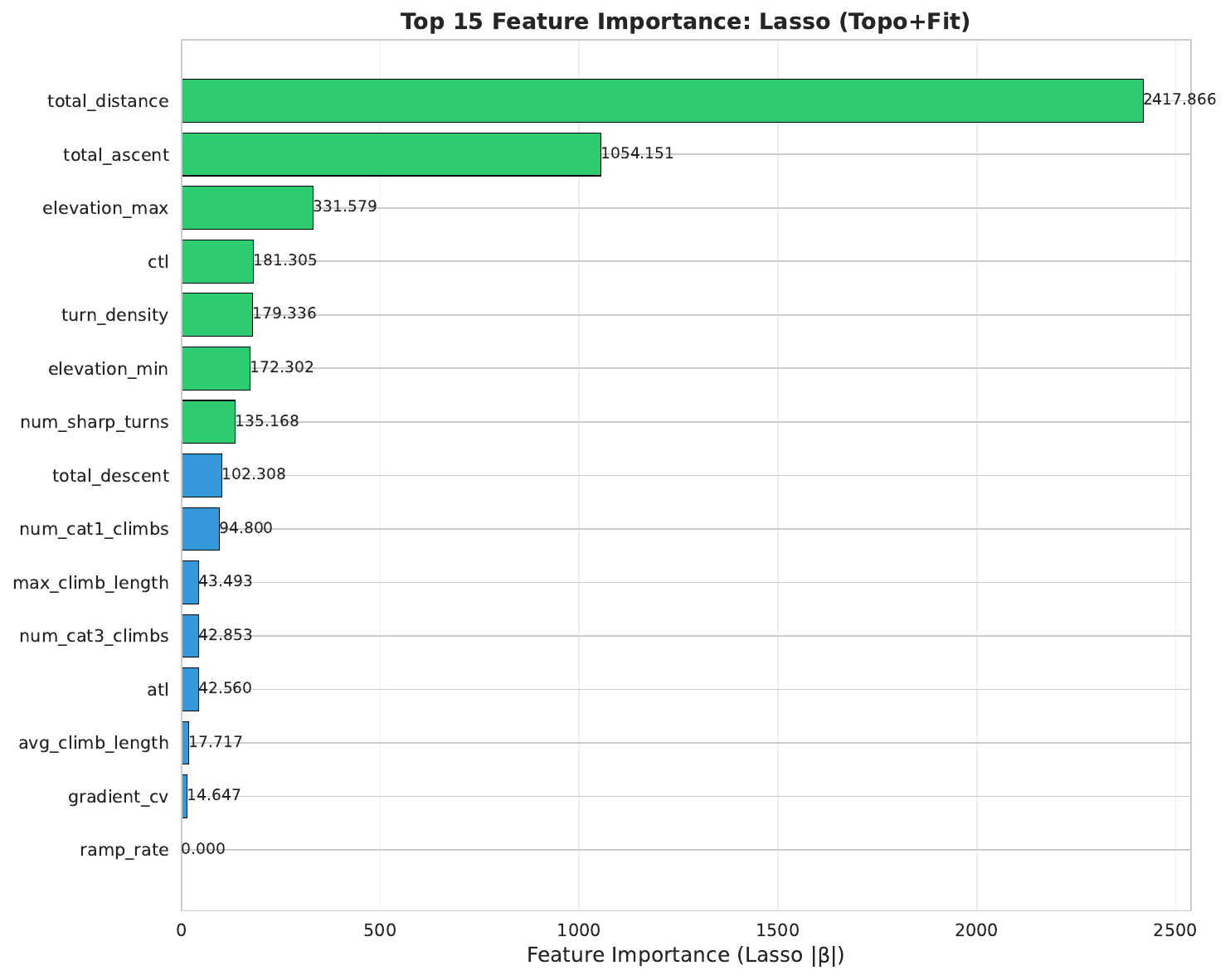}
\caption{SHAP feature importance for Lasso (Topology) model. Bars show mean |SHAP value|
in minutes, indicating average contribution to prediction. Distance and elevation metrics
dominate, but terrain complexity features (punchiness, climb density) contribute 25-30\%
of predictive power.}
\label{fig:shap_importance}
\end{figure}

Figure~\ref{fig:shap_importance} shows mean absolute SHAP values, quantifying each
feature's average contribution to predictions. The top 5 features account for over
80\% of predictive power:

\begin{enumerate}
\item \textbf{Distance} (22.5 min): Primary driver of duration
\item \textbf{Total Ascent} (16.5 min): Cumulative vertical meters add time non-linearly
\item \textbf{Elevation Gain/km} (12.5 min): Normalized difficulty metric (m/km)
\item \textbf{\% Grade 4-6\%} (5.1 min): Percent of route in moderate climbing zone
\item \textbf{Punchiness Score} (4.1 min): Terrain variability affects pacing
\end{enumerate}

\subsubsection{Feature Value Effects}
\begin{figure}[H]
\centering
\includegraphics[width=0.85\textwidth]{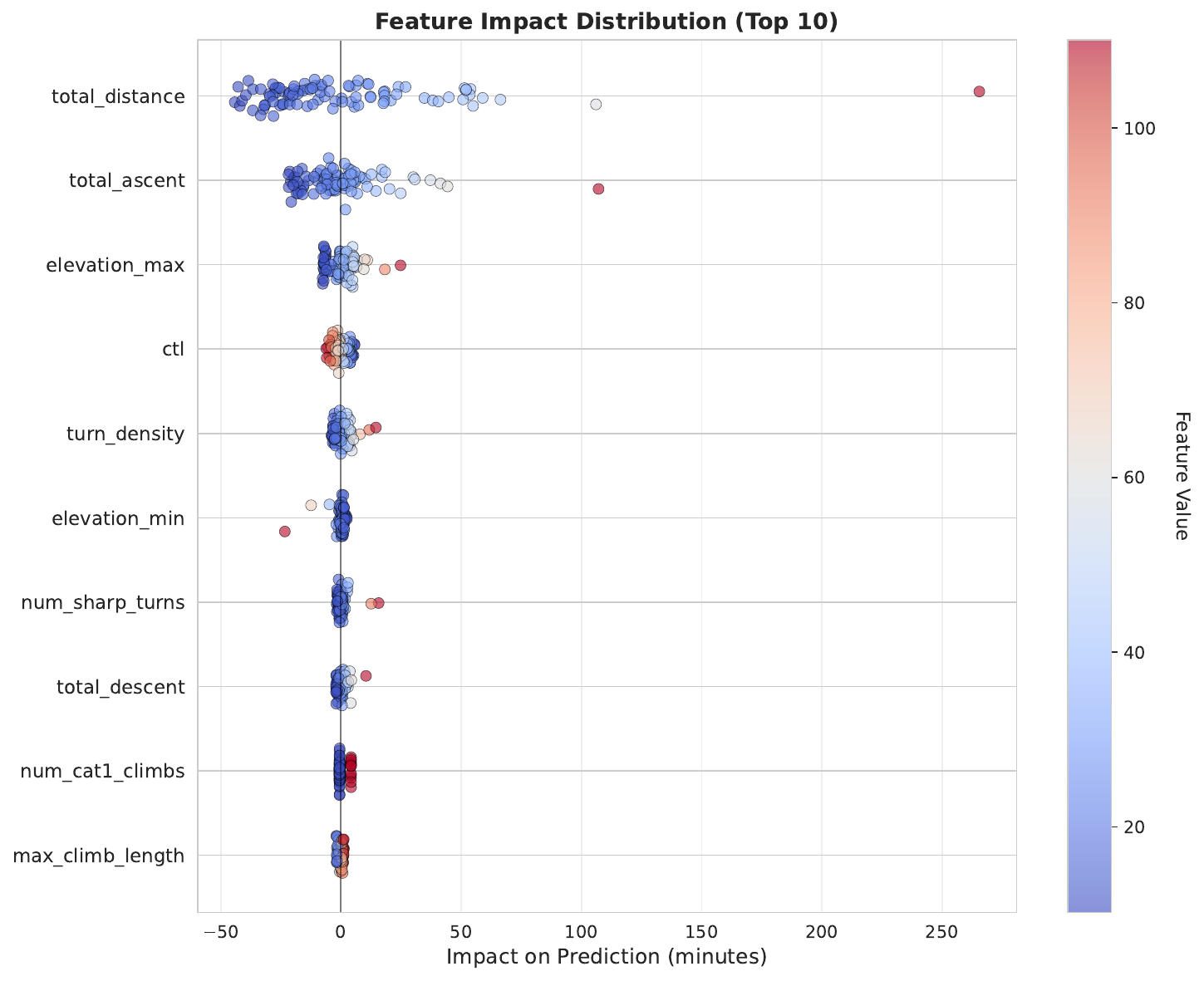}
\caption{SHAP summary plot showing feature value effects. Each point represents one
activity, colored by feature value (red=high, blue=low). Horizontal position shows
SHAP value (impact on prediction). Distance and ascent show strong positive
effects; punchiness adds time for high values.}
\label{fig:shap_summary}
\end{figure}

Figure~\ref{fig:shap_summary} visualizes how feature values influence predictions:
Distance and ascent show strong linear positive relationships (higher values $\rightarrow$
longer time), while punchiness score exhibits non-linear effects---high punchiness
(rolling terrain) adds 5-10 minutes even at equal distance/elevation, validating the
importance of gradient variability in duration prediction.

\section{Application: Track 101 MTB Case Study}

Having validated the model's accuracy, we now demonstrate its practical application
on a real-world route planning scenario. We evaluate the model on Track 101 MTB 2025,
a challenging $\sim$101 km mountainous route with 3,279m elevation gain (32.5 m/km difficulty
index, 1.5x dataset mean).

\subsection{Progressive Checkpoint Predictions}

Predictions are made at 25\%, 50\%, 75\%, and 100\% distance checkpoints
using Lasso (Topology) trained on the full 96-activity dataset.

\paragraph{Prediction Methodology}
At each checkpoint, the model predicts \textit{total route duration} using topology
features extracted from the GPX up to that point. At 25\% completion, only the first
quarter of the route is analyzed; at 100\%, the complete profile is used. This reveals
how duration estimates evolve as terrain difficulty becomes apparent.

The model uses only the \textbf{27 topology features} extracted from the truncated
GPX file (start to checkpoint). No fitness state, no historical data. This simulates
pre-ride planning where an athlete progressively examines route segments.

For Track 101 MTB 2025, we created checkpoint GPX files at 25\%, 50\%, 75\%, and
100\% distance and predicted total duration at each stage. The substantial increase
from 201 min (25\%) to 648 min (100\%) reveals this route's back-loaded difficulty---the
first quarter appears manageable, but cumulative climbing in later sections dramatically
extends the predicted finish time.

\subsubsection{Prediction Evolution}
Table~\ref{tab:checkpoint_preds} shows predictions evolve substantially as terrain is
revealed. The prediction more than triples from 25\% completion (201 min) to 100\%
(648 min), indicating this route's difficulty is heavily back-loaded.

\begin{table}[H]
\centering
\small
\begin{tabular}{@{}l r r r r@{}}
\hline
\textbf{Checkpoint} & \textbf{Dist (km)} & \textbf{Elev (m)} & \textbf{Climbs} & \textbf{Predicted Time (min)} \\
\hline
25\% & 25.2 & 671 & 5 & 200.9 \\
50\% & 50.4 & 1,023 & 8 & 293.6 \\
75\% & 75.6 & 1,939 & 13 & 450.7 \\
100\% & 100.7 & 3,279 & 18 & 647.6 \\
\hline
\end{tabular}
\caption{Progressive checkpoint predictions for Track 101 MTB 2025. Prediction
more than triples from 25\% to 100\% due to back-loaded climbing (only 5 climbs
in first quarter, 18 total).}
\label{tab:checkpoint_preds}
\end{table}

\begin{figure}[H]
\centering
\includegraphics[width=\textwidth]{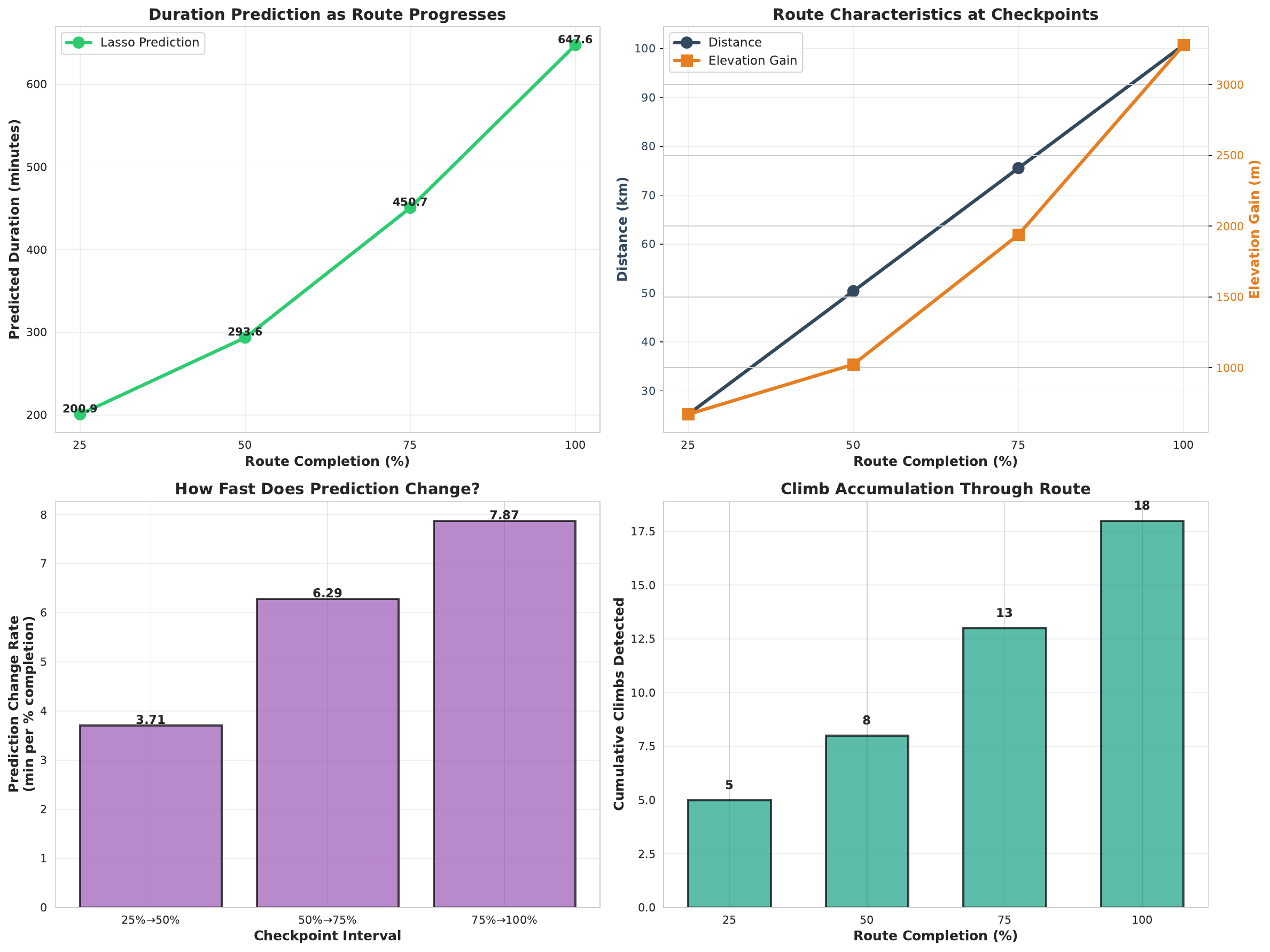}
\caption{Progressive checkpoint predictions for Track 101 MTB 2025. Top-left: Predicted
duration more than triples from 25\% to 100\% completion. Top-right: Distance and elevation
accumulation through route. Bottom-left: Prediction change rate accelerates in final
quarter (7.87 min/\% vs. 3.71 min/\% in first quarter). Bottom-right: Only 5 climbs
detected at 25\%, revealing back-loaded difficulty distribution.}
\label{fig:gpx_checkpoint}
\end{figure}

\subsubsection{Route Profile and Checkpoint Visualization}
\begin{figure}[H]
\centering
\includegraphics[width=\textwidth]{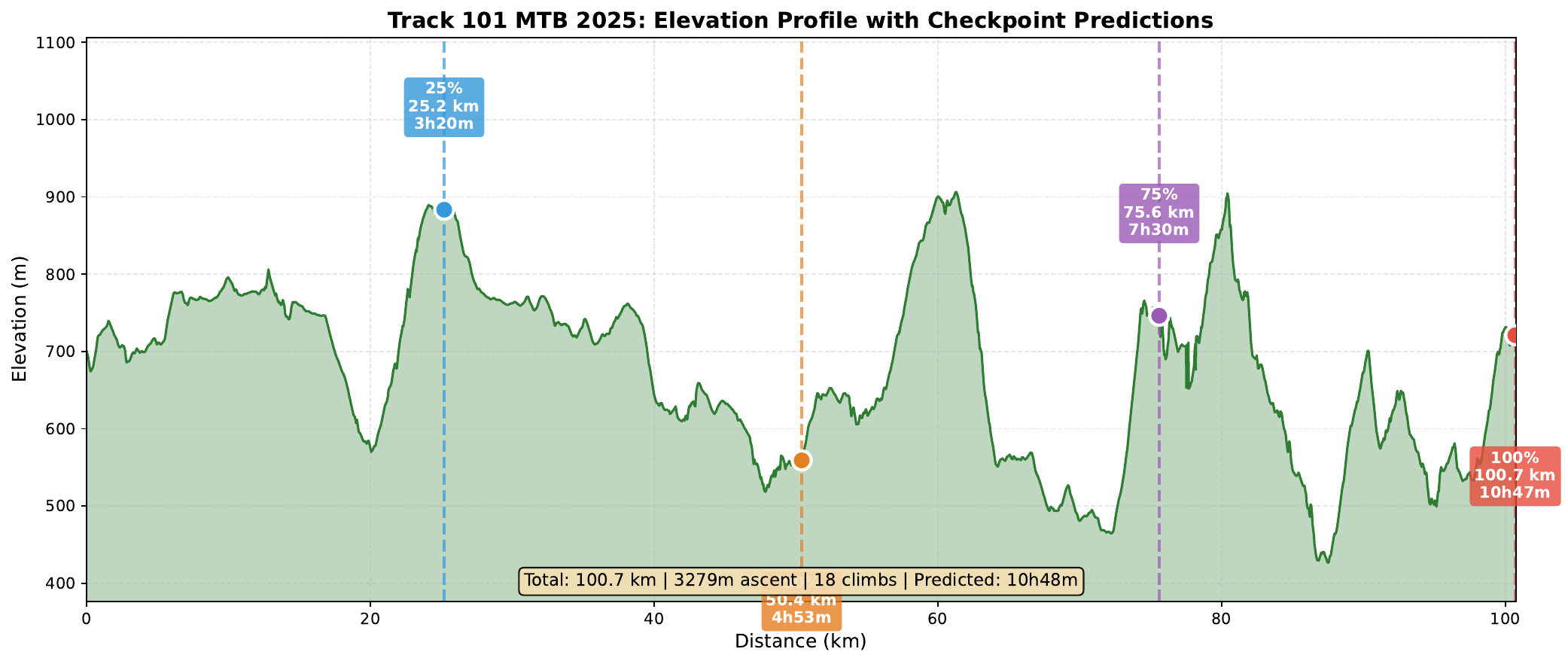}
\caption{Track 101 MTB 2025 elevation profile with progressive checkpoint predictions.
Vertical lines mark 25\%, 50\%, 75\%, and 100\% distance checkpoints. Each annotation
shows cumulative distance and predicted duration (e.g., 3h20m at 25\%, 10h47m at finish).
The profile reveals back-loaded climbing: gentle rolling terrain until km 50, followed
by sustained climbs in the final half. Total route: 100.7 km, 3,279m ascent, 18 climbs.}
\label{fig:gpx_elevation_profile}
\end{figure}

Figure~\ref{fig:gpx_elevation_profile} shows the elevation profile of Track 101 MTB 2025
with checkpoint locations and predicted durations marked. The profile reveals the
back-loaded difficulty: relatively gradual terrain in the first half gives way to
sustained climbing in the final 50km, with major ascents concentrated between km 60--90.

\subsubsection{Back-Loaded Route Difficulty}
The route accumulates 18 climbs total, but only 5 are visible at 25\% completion
(Figure~\ref{fig:gpx_checkpoint}). Prediction change rate accelerates from 3.71 min/\%
(25-50\%) to 7.87 min/\% (75-100\%), indicating steeper, longer climbs concentrate in
the final quarter. With a difficulty index of 32.5 m/km (vs. dataset mean of 21.0 m/km),
this route represents a ``very challenging'' classification---critical information
for nutrition and pacing strategy.

\subsection{ClimbPro Analysis}

\begin{figure}[H]
\centering
\includegraphics[width=\textwidth]{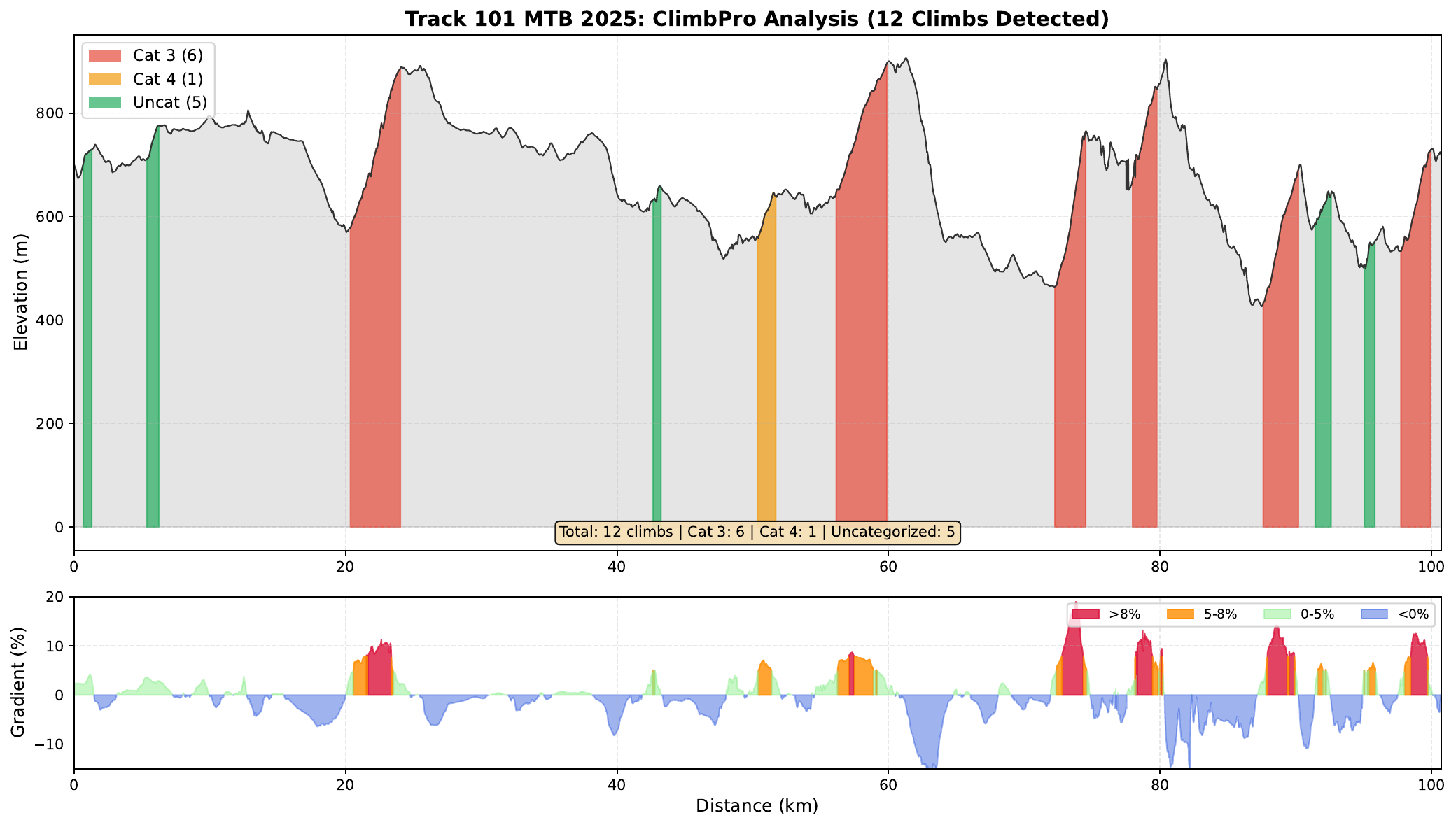}
\caption{ClimbPro analysis for Track 101 MTB 2025. Top: Elevation profile with 18
detected climbs highlighted by category (Cat 3 in orange, Cat 4 in yellow, uncategorized
in green). Bottom: Gradient distribution showing sections above 8\% (red), 5--8\%
(orange), and below 5\% (green). The route contains 6 Cat 3 climbs concentrated in
the final 40km.}
\label{fig:gpx_climbpro}
\end{figure}

Applying the Garmin ClimbPro algorithm to Track 101 MTB reveals how climb difficulty
accumulates across checkpoints. Figure~\ref{fig:gpx_climbpro} shows the elevation
profile with all detected climbs highlighted by category, and the gradient distribution
below. Table~\ref{tab:climbpro_analysis} shows the progressive detection of climbs
as the route unfolds.

\begin{table}[H]
\centering
\small
\begin{tabular}{@{}l r r r r r r@{}}
\hline
\textbf{Checkpoint} & \textbf{Climbs} & \textbf{Cat 3} & \textbf{Cat 4} & \textbf{Score} & \textbf{Max 500m Grad} & \textbf{Longest (km)} \\
\hline
25\% & 5 & 1 & 0 & 47,244 & 12.02\% & 3.83 \\
50\% & 8 & 1 & 0 & 61,117 & 12.02\% & 3.83 \\
75\% & 13 & 3 & 1 & 130,740 & 18.93\% & 6.01 \\
100\% & 18 & 6 & 1 & 215,879 & 19.13\% & 6.01 \\
\hline
\end{tabular}
\caption{ClimbPro analysis for Track 101 MTB 2025. Climb count and category distribution
reveal back-loaded difficulty: 72\% of total climb score (156k of 216k) concentrates in
the final 50\% of the route. Cat 3 climbs increase from 1 to 6 between 50\% and 100\%.}
\label{tab:climbpro_analysis}
\end{table}

The ClimbPro scoring ($S = d \times g$) quantifies this progression: at 50\% distance,
only 28\% of total climb difficulty has been encountered (61k of 216k score). The
sudden jump at 75\% (131k score) signals the route's demanding final quarter.

\subsection{Maximum Sustained Gradient Analysis}

\begin{figure}[H]
\centering
\includegraphics[width=\textwidth]{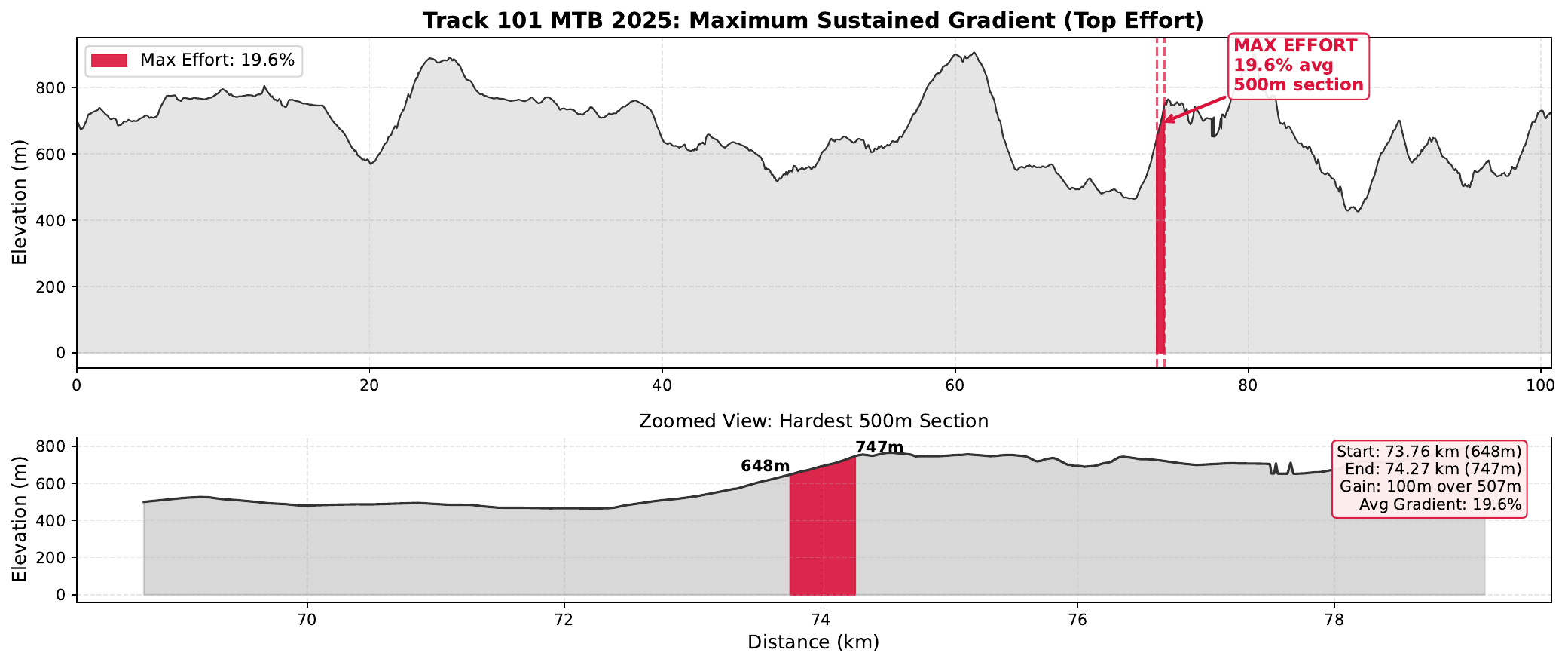}
\caption{Maximum sustained gradient analysis for Track 101 MTB 2025. Top: Full route
with the hardest 500m section highlighted in red (19.6\% average gradient at km 73.8--74.3).
Bottom: Zoomed view of the max effort section showing 99m elevation gain over 500m.
This metric captures peak physiological demand---the single hardest sustained effort
regardless of total elevation.}
\label{fig:gpx_max_effort}
\end{figure}

The \textit{max sustained gradient over 500m} metric identifies the single hardest
sustained effort on the route---the 500-meter section with highest average gradient.
Figure~\ref{fig:gpx_max_effort} highlights this section on Track 101 MTB.

For Track 101 MTB, the maximum sustained gradient is \textbf{19.6\%} located at km 73.8--74.3,
within the demanding final quarter. The evolution across checkpoints (12.0\% at 25\%,
12.0\% at 50\%, 18.9\% at 75\%, 19.6\% at 100\%) confirms the back-loaded difficulty.

This metric captures peak physiological demand beyond what aggregate statistics reveal.
A route with identical total elevation could have maximum sustained gradients of 8\%
(steady) or 20\% (punchy)---requiring fundamentally different pacing strategies.
The 19.6\% maximum on Track 101 indicates sections requiring near-maximal effort,
even for well-trained athletes, and validates the ``very challenging'' classification.

\paragraph{Practical Application}
Progressive predictions enable race-day decisions: at 50\% completion (50.4 km),
the model predicts 293.6 min total, revealing that the second half will take
significantly longer than the first. Athletes can adjust effort accordingly,
conserving energy for the demanding final quarter. This ``checkpoint intelligence''
transforms static pre-race duration estimates into dynamic race-day tools.

\subsection{Fitness Feature Contribution}

The Lasso (Topology + Fitness) model achieves 14\% lower MAE than topology alone,
confirming that fitness state contributes predictive value. However, interpreting
individual fitness coefficients requires caution: with N=96 and correlated predictors,
coefficient directions may reflect training data patterns (e.g., fitter athletes
attempting harder routes) rather than isolated causal effects.

For Track 101, varying fitness parameters while holding topology constant yields
predictions within a 5-minute range, confirming that \textbf{route topology dominates
duration prediction}. The minimum predicted duration remains $\sim$10h50m regardless
of fitness assumptions, revealing that the 32.5 m/km difficulty index and 18 detected
climbs fundamentally constrain achievable speed.

The practical implication is clear: for challenging routes like Track 101, athletes
should prioritize route-specific preparation---pacing strategy, nutrition planning,
and checkpoint targeting---rather than expecting large time gains from fitness
optimization alone.

\section{Discussion and Conclusion}

\subsection{Discussion}

\subsubsection{Topology Sets the Baseline, Fitness Refines the Prediction}
Route topology is the primary driver of duration, explaining over 88\% of the variance
in our dataset. Features like \textit{punchiness} (gradient variability) capture
difficulty beyond simple elevation gain~\cite{leo2022durability}. However, the
integration of fitness metrics (CTL, ATL) provides a critical 14\% improvement in
accuracy (MAE reduced from 7.66 to 6.60 min).

This finding suggests that while athletes may self-pace to manage effort, their
``cruising speed'' is statistically associated with their chronic training load (CTL) and
acute fatigue state (ATL). The 14\% error reduction (MAE improved by $\approx$1 minute)
may seem modest, but compounds meaningfully over training cycles: more accurate predictions
enable better pacing strategy and event preparation.

\subsubsection{Fitness Improves Prediction, but Interpretation Requires Caution}
While fitness features improve predictive accuracy by 14\%, interpreting individual
coefficients for ``what-if'' scenario analysis requires caution. With N=96 and
correlated predictors, learned coefficients may reflect confounding patterns in the
training data---fitter athletes tend to attempt harder routes---rather than isolated
causal effects. For example, the model's fitness coefficients cannot reliably answer
questions like ``how much faster will I be if I taper before race day?''

Despite this limitation, the aggregate contribution of fitness features is valid:
including CTL, ATL, and related metrics consistently reduce prediction error across
cross-validation folds. Future work with larger datasets or experimental designs
(e.g., repeated measurements under controlled fitness variations) could enable
more reliable causal inference for training periodization guidance.

\subsubsection{The Importance of Rigorous Feature Selection}
Our evaluation highlights the risk of data leakage in sports analytics. Initial
iterations using VAM (Vertical Ascent Speed) yielded artificially high accuracy
(R²$>$0.97) but failed to generalize for future predictions. By strictly isolating
\textit{a priori} features (Topology and Fitness) from \textit{a posteriori} results
(Speed, VAM), we achieved a realistic and deployable model (R²=0.922).

\subsubsection{Regularization Critical for Small Datasets}
With N=96 and 27-79 features, overfitting risk is severe. Random Forest, despite
aggressive Bayesian hyperparameter tuning via Optuna (30 trials per fold), achieves
only MAE=13.36---worse than simple baselines---with catastrophic overfitting.
Lasso's L1 penalty proves essential, producing stable predictions and automatic
feature selection. This negative result has practical value: practitioners should
default to regularized linear models for small sports performance datasets until
N $>$ 500.

\subsubsection{N-of-1 Study Design Justification}
Single-athlete studies are appropriate when: (1) \textbf{High inter-individual
variability}: Cycling performance depends on Functional Threshold Power, VO$_2$max, lactate threshold, W',
body composition, and skill---parameters varying 2-5x across amateur cyclists.
Population models would require extensive physiological testing (lab VO$_2$max, power
profiling) impractical for the target use case. (2) \textbf{Personalized prediction
goal}: This work targets individual athletes seeking ``How long will this route take
ME?'' rather than ``How long for an average cyclist?'' Personalized models learn
athlete-specific strengths (climbing vs. flat speed) and pacing tendencies.
(3) \textbf{Data availability}: Modern cyclists generate thousands of GPS activities,
providing abundant within-subject data without requiring sparse multi-athlete cohorts.

Similar N-of-1 designs have proven effective in chronobiology, nutrition, and clinical
medicine~\cite{kravitz2014design}. The key is sufficient within-subject variation in
predictor variables (route characteristics) and outcome (duration), which our dataset
provides.

\subsection{Limitations}

\paragraph{Dataset Size}
While N=96 is adequate for Lasso regression (learning curve plateaus at $\sim$60
samples), it precludes exploration of deep learning models. Gradient boosting or
neural networks might capture non-linearities inaccessible to linear models, but
require N $>$ 500 for stable training on this feature dimensionality.

\paragraph{Single Athlete}
Results may not generalize to cyclists with different physiology, skill levels, or
pacing strategies. A competitive racer might show strong fitness-performance coupling
(no self-pacing), while a novice might exhibit different duration-gradient relationships.
The model is intentionally personalized, requiring individual training data.

\paragraph{Environmental Factors Excluded}
Weather conditions (wind, temperature, precipitation), road surface quality, and
traffic are not captured. Wind alone can alter duration by 10-30\% on exposed routes.
GPS data lack environmental sensors; integrating weather API data are future work.

\paragraph{Pacing Assumptions}
The model assumes consistent pacing strategy (self-selected for training, maximal for
races). Strategic pacing variations (conserving energy for final climb) are not modeled.
Duration prediction implicitly averages over typical pacing behavior learned from
historical data.

\paragraph{GPS Accuracy}
Elevation data from barometric altimeters have $\pm$5m precision, while GPS elevation
can drift $\pm$20m. Systematic errors in climb detection or gradient calculation could
propagate to predictions, though high-quality devices mitigate this limitation.

\subsection{Future Work}

\paragraph{Multi-Athlete Validation}
Expanding to N=10-50 athletes would enable population-level models and quantify
personalization benefit magnitude. Transfer learning could pre-train on population
data, then fine-tune to individuals with limited personal history, reducing cold-start
requirements for new users.

\paragraph{Environmental Integration}
Weather API integration (wind speed/direction, temperature) would improve race-day
prediction accuracy. Wind affects duration by 10-30\% on exposed routes; historical
weather data aligned to GPS tracks could train weather-aware models for improved
real-world applicability.

\paragraph{Real-Time Pacing Guidance}
Current work predicts total duration; real-time systems could provide live feedback:
``You're 2 min ahead of predicted pace, ease off to avoid bonking on final climb.''
This requires modeling remaining time given current position and accumulated fatigue,
an open research problem in endurance sports analytics.

\paragraph{Segment-Level Predictions}
Rather than total duration, predicting time for individual climbs or segments enables
Strava segment performance forecasting and climb-specific race preparation. This
requires segment-level feature extraction and residual fatigue modeling across
cumulative efforts.

\subsection{Conclusion}

This work demonstrates that machine learning models can predict cycling duration from
route topology and fitness state with high accuracy (MAE=6.60 minutes, R²=0.922) using
only consumer-grade GPS data and historical training load. Unlike physics-based approaches
requiring aerodynamic testing and real-time wind forecasts, our method learns athlete-specific
performance patterns from historical rides, making personalized predictions accessible to
amateur cyclists.

Key contributions include: (1) \textbf{Novel terrain features} grounded in exercise
physiology (punchiness, climb detection matching Garmin ClimbPro, gradient distribution),
(2) \textbf{Rigorous feature selection} eliminating data leakage from result-based metrics
(VAM, speed, power), ensuring deployment validity, (3) \textbf{Regularization importance}
demonstrated through Random Forest overfitting on N=96, validating Lasso's robustness for
small sports datasets, and (4) \textbf{Fitness integration} showing training load metrics
(CTL, ATL) provide 14\% accuracy improvement over topology alone, though individual
coefficient interpretation requires caution due to predictor correlation.

The N-of-1 study design proves viable for personalized sports analytics, leveraging
abundant within-subject GPS data to overcome inter-individual variability challenges.
Progressive checkpoint predictions enable race planning, though back-loaded route
difficulty limits early estimate stability.

This approach enables practical applications without specialized equipment: pre-ride
duration estimates for training planning, race pacing strategy, and route difficulty
assessment. The work showcases an end-to-end machine learning pipeline suitable for
demonstrating data collection, rigorous feature engineering, proper validation, and
deployment-ready prediction capabilities.

\subsection{Data and Code Availability}

The feature extraction pipeline, model training code, and analysis notebooks are available
at \url{https://github.com/fran-aguila/health-hub}. Due to privacy considerations, the
raw GPS activity data and training logs are not publicly released. Researchers interested
in replicating this work can apply the methodology to their own cycling data using standard
export formats from platforms such as Strava, Garmin Connect, or Intervals.icu.

\printbibliography

@article{banister1991modeling,
  author = {Banister, Eric W.},
  title = {Modeling Elite Athletic Performance},
  journal = {Physiological Testing of Elite Athletes},
  pages = {403--424},
  year = {1991},
  publisher = {Human Kinetics}
}

@article{sanders2017methods,
  author = {Sanders, Dajo and Heijboer, Mathieu},
  title = {Methods of Monitoring Training Load and Their Relationships to Changes in Fitness and Performance in Competitive Road Cyclists},
  journal = {International Journal of Sports Physiology and Performance},
  volume = {12},
  number = {5},
  pages = {668--675},
  year = {2017},
  doi = {10.1123/ijspp.2016-0454}
}

@article{seiler2006quantifying,
  author = {Seiler, Stephen and Kjerland, Glenn Øvrevik},
  title = {Quantifying Training Intensity Distribution in Elite Endurance Athletes: Is There Evidence for an ``Optimal'' Distribution?},
  journal = {Scandinavian Journal of Medicine \& Science in Sports},
  volume = {16},
  number = {1},
  pages = {49--56},
  year = {2006},
  doi = {10.1111/j.1600-0838.2004.00418.x}
}

@article{leo2022power,
  author = {Leo, Peter and Spragg, James and Podlogar, Tim and Lawley, Justin S. and Mujika, Iñigo},
  title = {Power Profiling and the Power-Duration Relationship in Cycling: A Narrative Review},
  journal = {European Journal of Applied Physiology},
  volume = {122},
  pages = {301--316},
  year = {2022},
  doi = {10.1007/s00421-021-04833-y}
}

@article{leo2022durability,
  author = {Leo, Peter and Spragg, James and Mujika, Iñigo and Menz, Verena and Lawley, Justin S.},
  title = {Durability in Elite Endurance Athletes},
  journal = {Scandinavian Journal of Medicine \& Science in Sports},
  volume = {32},
  pages = {1565--1578},
  year = {2022},
  doi = {10.1111/sms.14699}
}

@article{skiba2015intramuscular,
  author = {Skiba, Philip F. and Fulford, Jonathan and Clarke, David C. and Vanhatalo, Anni and Jones, Andrew M.},
  title = {Intramuscular Determinants of the Ability to Recover Work Capacity above Critical Power},
  journal = {European Journal of Applied Physiology},
  volume = {115},
  pages = {703--713},
  year = {2015},
  doi = {10.1007/s00421-014-3050-3}
}

@article{abbiss2008describing,
  author = {Abbiss, Chris R. and Laursen, Paul B.},
  title = {Describing and Understanding Pacing Strategies during Athletic Competition},
  journal = {Sports Medicine},
  volume = {38},
  number = {3},
  pages = {239--252},
  year = {2008},
  doi = {10.2165/00007256-200838030-00004}
}

@article{padilla2000scientific,
  author = {Padilla, Sabino and Mujika, Iñigo and Orbañanos, Jon and Angulo, Francisco},
  title = {Exercise Intensity during Competition Time Trials in Professional Road Cycling},
  journal = {Medicine \& Science in Sports \& Exercise},
  volume = {32},
  number = {4},
  pages = {850--856},
  year = {2000},
  doi = {10.1097/00005768-200004000-00019}
}

@article{martin1998validation,
  author = {Martin, James C. and Milliken, Douglas L. and Cobb, John E. and McFadden, Kerry L. and Coggan, Andrew R.},
  title = {Validation of a Mathematical Model for Road Cycling Power},
  journal = {Journal of Applied Biomechanics},
  volume = {14},
  number = {3},
  pages = {276--291},
  year = {1998},
  doi = {10.1123/jab.14.3.276}
}

@book{coggan2019tss,
  author = {Allen, Hunter and Coggan, Andrew R.},
  title = {Training and Racing with a Power Meter},
  publisher = {VeloPress},
  edition = {3rd},
  year = {2019},
  note = {Foundational text for TSS, NP, IF metrics}
}

@article{jobson2009prediction,
  author = {Jobson, Simon A. and Passfield, Louis and Atkinson, Greg and Barton, Gabriela and Scarf, Philip},
  title = {The Analysis and Utilization of Cycling Training Data},
  journal = {Sports Medicine},
  volume = {39},
  number = {10},
  pages = {833--844},
  year = {2009},
  doi = {10.2165/11317840-000000000-00000}
}

@article{menaspa2017cycling,
  author = {Menaspà, Paolo and Quod, Marc and Martin, David T. and Peiffer, Jeremiah J. and Abbiss, Chris R.},
  title = {Physical Demands of Sprinting in Professional Road Cycling},
  journal = {International Journal of Sports Medicine},
  volume = {38},
  number = {13},
  pages = {1003--1008},
  year = {2017},
  doi = {10.1055/s-0043-114007}
}

@article{chen2020machine,
  author = {Chen, Xiang and Zhang, Yao and Wang, Zhi-Yong and Li, Shi-Qi},
  title = {A Review of Machine Learning Methods for Sports Outcome Prediction},
  journal = {Applied Sciences},
  volume = {10},
  number = {22},
  pages = {8048},
  year = {2020},
  doi = {10.3390/app10228048}
}

@article{diprampero1979equation,
  author = {di Prampero, Pietro E. and Cortili, G. and Mognoni, P. and Saibene, F.},
  title = {Equation of Motion of a Cyclist},
  journal = {Journal of Applied Physiology},
  volume = {47},
  number = {1},
  pages = {201--206},
  year = {1979},
  doi = {10.1152/jappl.1979.47.1.201}
}

@inproceedings{akiba2019optuna,
  author = {Akiba, Takuya and Sano, Shotaro and Yanase, Toshihiko and Ohta, Takeru and Koyama, Masanori},
  title = {Optuna: A Next-generation Hyperparameter Optimization Framework},
  booktitle = {Proceedings of the 25th ACM SIGKDD International Conference on Knowledge Discovery \& Data Mining},
  year = {2019},
  pages = {2623--2631},
  doi = {10.1145/3292500.3330701}
}

@inproceedings{lundberg2017unified,
  author = {Lundberg, Scott M. and Lee, Su-In},
  title = {A Unified Approach to Interpreting Model Predictions},
  booktitle = {Advances in Neural Information Processing Systems},
  volume = {30},
  year = {2017},
  pages = {4765--4774}
}

@article{kravitz2014design,
  author = {Kravitz, Richard L. and Duan, Naihua and {DEcIDE Methods Center N-of-1 Guidance Panel}},
  title = {Design and Implementation of N-of-1 Trials: A User's Guide},
  journal = {AHRQ Publication No. 13(14)-EHC122-EF},
  year = {2014},
  publisher = {Agency for Healthcare Research and Quality}
}

\end{document}